\documentclass{article}

\usepackage{PRIMEarxiv}

\usepackage[utf8]{inputenc} 
\usepackage[T1]{fontenc}    
\usepackage{hyperref}       
\usepackage{url}            
\usepackage{booktabs}       
\usepackage{amsfonts}       
\usepackage{nicefrac}       
\usepackage{microtype}      
\usepackage{fancyhdr}       
\usepackage{graphicx}       
\graphicspath{{media/}}     

\usepackage{amsmath,amssymb,amsfonts,amsthm}
\usepackage{color,soul}

\title{An entropy-optimal path to humble AI}

\author{
  Davide Bassetti\thanks{These authors contributed equally to this work.}\\
  Faculty of Mathematics\\
  Rheinland-Pf\"{a}lzische Technische Universit\"{a}t Kaiserslautern Landau,\\
  Kaiserslautern, Germany\\
  \\
  \And
  Luk\'{a}\v{s} Posp\'{i}\v{s}il$^*$ \\
  Department of Mathematics\\
  Faculty of Civil Engineering\\
  VSB - Technical University of Ostrava\\
  Ostrava, Czech Republic\\
  \And
  Michael Groom \\
  CSIRO Environment, \\
  Eveleigh, New South Wales, Australia
  \And
  Terence J. O'Kane\\
  CSIRO Environment, \\
  Battery Point, Tasmania, Australia\\
  \And
  Illia Horenko\thanks{To whom correspondence should be addressed, at \texttt{horenko@rptu.de}}\\
  Faculty of Mathematics\\
  Rheinland-Pf\"{a}lzische Technische Universit\"{a}t Kaiserslautern Landau,\\
  Kaiserslautern, Germany\\
  \texttt{horenko@rptu.de}\\
}

\usepackage[dvipsnames]{xcolor}

\usepackage{float}
\usepackage[ruled]{algorithm2e}
\SetKwComment{Comment}{/* }{ */} 

\DeclareMathAlphabet{\mymathbb}{U}{BOONDOX-ds}{m}{n} 
\newcommand{\onebb}{\mymathbb{1}} 
\newcommand{\zerobb}{\mymathbb{0}} 
\DeclareMathOperator*{\diag}{diag}

\newcommand{\softmax}{\mathrm{softmax}}
\newcommand{\iit}{\mathrm{it}} 
\newcommand{\Ksum}{K_{\mathrm{sum}}}

\newcommand{\sn}[2]{$#1 \times 10^{#2}$}

\usepackage{ulem} 

\usepackage{thmtools}
\newtheorem{theorem}{Theorem}[section]
\usepackage{mdframed} 
\newmdtheoremenv[
  backgroundcolor=white,
  linecolor=gray!50,
  linewidth=1pt,
  innertopmargin=5pt,
  innerbottommargin=5pt,
  innerrightmargin=5pt,
  innerleftmargin=5pt,
  roundcorner=0pt,
  skipabove=20pt,
  skipbelow=20pt,
]{lemma}[theorem]{Lemma}

\begin{document}
\maketitle

\begin{abstract}
  Progress of AI has led to very successful, but by no means humble models and tools, especially regarding (i)  the huge and further exploding costs and resources they demand, and (ii) the over-confidence of these tools with the answers they provide.  Here we introduce a novel mathematical framework for a non-equilibrium entropy-optimizing reformulation of Boltzmann machines based on the exact law of total probability and the exact convex polytope representations. We show that it results in the highly-performant, but much cheaper, gradient-descent-free learning framework with mathematically-justified existence and uniqueness criteria, and cheaply-computable confidence/reliability measures for both the model inputs and the outputs. Comparisons to state-of-the-art AI tools in terms of performance, cost and the model descriptor lengths on a broad set of synthetic and real-world problems with varying complexity reveal that the proposed method results in more performant and slim models, with the descriptor lengths being very close to the intrinsic complexity scaling bounds for the underlying problems. Applying this framework to historical climate data results in models with systematically higher prediction skills for the onsets of important  La Ni\~na and El Ni\~no climate phenomena, requiring just few years of climate data for training - a small fraction of what is necessary for contemporary climate prediction tools.
\end{abstract}


\section{Introduction}
A large proportion of modern neural networks are based on the perceptron feedforward network architecture \cite{rosenblattPerceptronProbabilisticModel1958a}, and their widespread success has been enabled by the introduction of the backpropagation algorithm, which provides a way to compute the gradient needed for applying gradient descent \cite{rumelhartLearningRepresentationsBackpropagating1986}.
In general, learning of modern AI models $M(\theta|X)$, with tuneable parameters $\theta$ and some given training data $X$, is formulated and performed as a numerical minimization of a loss function $L_M(\theta|X)$, which measures the quality of fitting the training data $X$ by evaluating how distant the output of the model $M$ is from the desired output.
Starting with some initial guess for the parameters $\theta^{(0)}$, the gradient descent method (GD) - and its stochastic variant SGD - attempt to iteratively move towards a (local) minimum of the loss $L_M(\theta|X)$, shifting the parameter iterates $\theta^{(i-1)}$ in the direction of the loss gradient $\nabla_{\theta}L_M(\theta^{(i-1)}|X)$, by computing updated values as $\theta^{(i)}=\theta^{(i-1)}-\alpha^{(i-1)}\nabla_{\theta}L_M(\theta^{(i-1)}|X)$.

The value of the learning rate $\alpha^{(i-1)}$ limits the size of the step in the direction of the gradient, and may either be fixed or varied according to a predefined schedule. Its selection is crucial for success and efficiency of both GD and SGD, since adopting larger values may result in large steps in the gradient direction and consequent ``overshooting'' of the minimum, whereas making $\alpha^{(i-1)}$ small will require more iterations for convergence and larger computational costs when training the model $M$. Based on the physics-inspired ideas introduced by B. Polyak and A. Nesterov - and interpreting $\theta^{(i)}$ and $\alpha^{(i)}$ as respectively a position and a momentum of a viscous Newtonian particle at time $(i)$ - allows defining equations for $\alpha^{(i)}$ as a discretisation of the stochastic Langevin equation for this virtual particle \cite{polyakIntroductionOptimization1987,nesterovIntroductoryLecturesConvex2013}.
These ideas have resulted in a plethora of more effective numerical algorithms than \emph{naive} GD and SGD, such as Nesterov's accelerated GD, momentum methods and algorithms like ADAM (for Adaptive Moment estimation, incorporating information about the second moment, or variance, of the gradients), which constitute the numerical backbone of modern AI \cite{bengioPracticalRecommendationsGradientBased2012, kingmaAdamMethodStochastic2017}.

Such advancements in the gradient descent procedure still present some drawbacks. Indeed, the steps in GD (and methods derived from it) are decided based on first order or at most approximated second order information. Thus, a large number of such steps may be required for convergence, making the training of those methods computationally expensive. Moreover, achieving satisfactory convergence relies on careful selection a number of user-defined parameters which need to be empirically chosen (for instance: batch size, learning rate schedule, etc.) and may depend on each other \cite{heControlBatchSize2019, kandelEffectBatchSize2020}. Employing more advanced learning algorithms may lead to the introduction of additional parameters to be tuned (e.g., momentum or decay rates for the moments in ADAM). Furthermore, the convergence also depends strongly on the smoothness properties of the loss function $L$ which are sensitive to the user-defined parameters, e.g., the type of nonlinearity (logistic, ReLU or other activation function variants) in model $M$, and may not be necessarily achieved \cite{reddiConvergenceAdam2019}.

Large amounts of data are typically required for training a performant neural network, leading to the widespread use of the expression ``big data'' to denote the type of problems where the application of deep learning has been particularly successful. In cases where the amount of data is instead limited and/or the data is high-dimensional, neural networks (as well as other AI tools) tend to manifest overfitting, i.e., memorization of the training data by the model without skill in generalization to new and yet unseen data \cite{yingOverviewOverfittingIts2019, horenkoScalableEntropicBreaching2020a}.

The size of AI models (and hence the dimension of the parameter vector $\theta$) underwent an exponential growth in the last years \cite{cottier2024rising}, and this for example reverberates in terms of inference cost for use of Large Language Models (LLMs) \cite{samsi2023words}. Despite this, recent results indicate issues in dealing with the intrinsic complexity inherent to the problems they face \cite{shojaee2025illusion}.

Furthermore, the decision functions of current state-of-the-art AI tools are defined on the entirety of the input space, which implies that any test data point respecting the condition of having the same shape as the input used for training will be mapped to a response, even if the network did not observe any similar data during the training. No confidence is generally reported on how far the test point is from the training data set. This can result in a lack of robustness and/or over-confidence, and has been illustrated by adversarial attacks that can flip the decisions of a neural network by minimally perturbing some data points that were used in training the model.

\subsection{Energy-based models}
An alternative to feedforward architectures trained using GD and related approaches are energy-based models, which will be briefly described in the following paragraphs. Drawing inspiration from neuroscience and statistical physics, Hopfield devised his eponymous networks with the goal of creating a system endowed with associative memory \cite{hopfieldNeuralNetworksPhysical1982}. The resulting system is a network composed of McCulloch-Pitts neurons with binary states, densely connected (with exclusion of self connections) using symmetric and bidirectional links. Each neuron is associated with a specific input of the network; for example, the presentation of a $n$ by $m$ image to the network requires a group of $n \times m$ neurons.

Given the similarity of the resulting system to a spin glass Ising model, Hopfield leveraged the mathematics from statistical physics to describe its behavior.
The Ising model recreates aspects of ferromagnetism using a collection of units in a lattice, where each unit possesses a spin (which can take one of two states) and direct neighbors are allowed to interact. If the spin of interacting units is the same value, the energy of the interaction will be lower than that of units of opposite spin. Thus, the total energy of the system can be calculated using the Hamiltonian function, which describes all interactions between neighbours, and the probability of a given spin configuration can be calculated using the Boltzmann distribution \cite{isingBeitragZurTheorie1925}. The energy landscape of an Ising model, and similarly of a Hopfield network, contains multiple stable local minima, but in the latter case their location can be influenced by the choice of weights used for connecting the neurons. Given a set of $T$ input memories from $\{-1,1\}^D$ collected in a matrix $X \in \mathbb{R}^{D\times T}$, the optimal weight matrix $J$ can be found as $J = XX^{\mathsf{T}}$.
Allowing the network to update the activations of the nodes $\gamma_i$ for $i=1,\dots,D$ (collected in a vector for convenience) from a given starting configuration to minimize the Hamiltonian of the system, given by the quadratic form

\begin{equation}\label{eqn:boltzmann}
  E = c^{\mathsf{T}} \gamma  - \gamma^{\mathsf{T}} J \gamma,
\end{equation}
where $c$ is a vector in $\mathbb{R}^D$, will result in convergence to local minima that correspond to memorized data, as long as the capacity allows. Such a network can store around $0.14 D$ memories, before presenting problems of overlap of local minima.
Note that the size of the network is constrained by the dimensionality of the data itself.

Boltzmann machines (BM) extend the Hopfield network by (i) introducing a group of neurons that is \emph{not} directly associated with an input, called hidden units to differentiate it from the visible units, and (ii) using stochastic units, with probabilistic sigmoidal activation functions. A simplified variant of this technique is the Restricted Boltzmann Machine (RBM) \cite{smolenskyChapter6Information1999, ackleyLearningAlgorithmBoltzmann1985}, which imposes absence of connections within layers, i.e., dense connectivity between the two groups but no direct connection between members of the same group. Employing such structure grants the advantage of being able to train the machines more efficiently \cite{hintonPracticalGuideTraining2012}. RBMs effectively learn probability distributions, by creating a model using the distribution of visible and hidden units, and do not require labels, allowing unsupervised learning as well as data generation (however, they can be also employed in classification tasks \cite{larochelleClassificationUsingDiscriminative2008}). Inference can be performed by fixing the visible units and allowing the hidden ones to reach a state of equilibrium.

Training RBMs is performed via contrastive divergence, a procedure involving two phases. Following initialisation using random weights, the following steps are repeated iterating over the training data: (i) Positive phase, involving the samples of the training set and (ii) negative phase, using some form of Gibbs sampling to approximate the negative gradient, followed by weight update, performed using a learning rate and the positive and negative gradients. The goal of the training is to obtain weights that result in the training examples being low energy states and other possible inputs being high energy states \cite{fischerTrainingRestrictedBoltzmann2014}, i.e., crafting an energy landscape where the minima are in favourable locations.
Several hidden layers can be stacked, forming Deep Belief Networks \cite{hintonLearningMultipleLayers2007a}. The key point to note is that the likelihood function is required to be normalised by a partition function that integrates over all possible configurations, the number of which grows exponentially with the number of units. Therefore, the gradients are in practice only approximated, resulting in slow convergence and expensive computations.

In conclusion, the training of neural and RBM networks may incur any of the above delineated problems, namely (i) it uses steps based upon first order or approximated higher order information, (ii) some level of configuration of the optimization procedure is required, (iii) a large amount of data is needed to achieve good performance and (iv) no guarantees are provided for whether a given test point is similar to the training examples or outside of the ``training region'' and whose prediction is therefore made with lower confidence.
Furthermore the training of RBMs involves the expensive approximation of the negative gradient due to its intractability, hence they have been superseded by other approaches for many tasks \cite{melkoRestrictedBoltzmannMachines2019}. Considering how these factors are a persistent challenge in training effective AI models, in the following section, we will propose a method that is devoid of such limitations.

\section{Results}\label{sec:results}

\subsection{Notation}
Throughout the manuscript, we will use the notation introduced in this paragraph.
We denote $\mathbb{R}^N$ as an $N$-dimensional real vector space and $\mathbb{R}^{M  \times N}$ as the vector space of real matrices of size $M \times N$. Throughout the equations, all vectors are assumed to be column vectors. The notation $v^{\mathsf{T}}$ denotes the transpose of the column vector $v \in \mathbb{R}^N$.
To shorten the notation, for any given matrix $A$, $A_{:,i}$ refers to the $i$th column whereas $A_{j,:}$ to the $j$th row.

The notation $\{ v \}_i$ represents the $i$-th component of the vector $v$, while $\{ A \}_{i,j}$ denotes the entry in the $i$-th row and $j$-th column of the matrix $A$. The mapping $\langle \cdot, \cdot \rangle: \mathbb{R}^N \times \mathbb{R}^N \mapsto \mathbb{R}$ represents the standard dot product. Specifically, for vectors $v, w \in \mathbb{R}^N$, we define
\begin{displaymath}
  \langle v, w \rangle := \sum\limits_{i = 1}^N \{ v \}_i \{ w \}_i.
\end{displaymath}

We denote $\mathbb{P}^K$ a set of discrete probability density vectors of size $K$ and $\mathbb{P}_L^{K_1 \times K_2}$ a set of all left-stochastic probability matrices, i.e.,
\begin{displaymath}
  \mathbb{P}^K :=
  \left\lbrace
  \gamma \in \mathbb{R}^K
  ~ \vert ~
  \forall k: \gamma_k \in [0,1]
  ~~\textrm{and}~
  \sum\limits_{k=1}^K \gamma_k = 1
  \right\rbrace,
  ~~~
  \mathbb{P}_L^{K_1 \times K_2} :=
  \left\lbrace
  \theta \in \mathbb{R}^{K_1 \times K_2}
  ~ \vert ~
  \forall k_2: \theta_{:,k_2} \in \mathbb{P}^{K_1}
  \right\rbrace.
\end{displaymath}

We define a component-wise logarithm $\log : \mathbb{R}^N \mapsto \mathbb{R}^N$ by $\lbrace \log w \rbrace_i = \log \lbrace w \rbrace_i$ and component-wise exponential function $\exp : \mathbb{R}^N \mapsto \mathbb{R}^N$ by $\lbrace \exp w \rbrace_i = \exp \lbrace w \rbrace_i$.

Matrix $I_K$ denotes identity matrix of size $K$, vector $\onebb_{K}$ denotes vector of ones of size $K$, vector $\zerobb_{K}$ is a vector of zeros of size $K$.

The function $\softmax : \mathbb{R}^N \mapsto \mathbb{P}^N$ is defined as
\begin{displaymath}
  \lbrace \softmax (x) \rbrace_i := \frac{\exp{x_i}}{\sum\limits_{j=1}^N \exp{x_j}}
\end{displaymath}
and the block softmax function
\begin{displaymath}
  \underset{K_1,\dots,K_N}{\softmax} :
  \left(
    \mathbb{R}^{K_1} \times \dots \times \mathbb{R}^{K_N}
  \right)
  \mapsto
  \left(
    \mathbb{P}^{K_1} \times \dots \times \in \mathbb{P}^{K_N}
  \right)
\end{displaymath}
is defined by
\begin{displaymath}
  \underset{K_1,\dots,K_N}{\softmax} \left(
    \left[
      \begin{array}{c}
        x_{(1)} \\
        \vdots \\
        x_{(N)}
      \end{array}
    \right]
  \right) :=
  \left[
    \begin{array}{c}
      \softmax (x_{(1)}) \\
      \vdots \\
      \softmax (x_{(N)})
    \end{array}
  \right]
  , ~~ \textrm{with blocks}~ x_{(n)} \in \mathbb{R}^{K_n}, n = 1,\dots,N.
\end{displaymath}

The mapping $\Vert . \Vert_2$ denotes the Euclidean vector norm or corresponding matrix norm. The mapping $\Vert . \Vert_F$ is Frobenius matrix norm.

\subsection{Formulation}\label{sec:formulation}
In this work, we present a methodology that extends classical Hopfield networks and RBMs, by allowing each state to be a probability distribution and deploying probability-measure-conforming distance measures in its formulation. We argue that the main bottlenecks in the existing formulations of BMs are \textit{(i) that deploying quadratic form (\ref{eqn:boltzmann}) for energy to probability measures $\gamma_i$ and $\gamma_j$ implies applying and minimizing the $J_{i,j}$-weighted squared Euclidean metric $J_{i,j}\gamma_i\gamma_j$  as a measure of distance, which is mathematically not an adequate measure of  distance between the probability distributions}, as well as (ii) the \textit{equilibrium assumption in the BM formulation}. The formulation that we will propose in the following introduces several advantages, first and foremost the possibility to operate on them using Bayesian tools, while additional considerations, such as an appropriate selection of a metric, significantly increases the power of the technique.

\paragraph{Inputs, outputs and architecture}
Let $X(1),\dots,X(T)$ be a sequence of $T$ $K_0$-dimensional data vectors, and $\pi(1),\dots,\pi(T)$ be a corresponding sequence of target values to be predicted (for example a continuous variable in case of regression or $K_{N+1}$-dimensional label probability distributions for classification). In contrast to Hopfield networks, real-valued input features are allowed and are adaptively discretized while fitting the model. Formulating the classification task using labels probability distributions extends and generalizes one-hot encoding, with the additional possibility to include information on confidence on the labels. For example, encoding a 90\% probability of data point $t$ to assume label 1 in a binary can be expressed as the corresponding $\pi(t) = [0.9, 0.1]$, and similarly, no information on datapoint $t$ in a  classification problem with $M$ possible classes can also be clearly defined as $\pi(t) = \left[\frac{1}{M},\dots,\frac{1}{M}\right] $.

The architecture of the model can be divided in two types of layer. The layers operate by transforming the input in a set of representations on probabilistic simplices, and in general try to minimize the distortion introduced by the use of their representation, as well as the reduction in performance with respect to the learning task.
For simplicity, we will limit this description to the general case of $N+1$ sequentially connected layers (i.e., in a feedforward way), but extensions to other architectures are straightforward.

\paragraph{Input layer}\label{par:input_layer}
The first layer (input layer) is different from all following ones, since it transforms the inputs to probability distributions, and measures the reliability of the inputs. It takes as input $X(t)$, and measures probabilities $\left\{\gamma_{(0)}\right\}_{d,t}$ of the relative importance of feature dimension $d$ for the instance $t$, comparing the inputs to a matrix $S\in\mathbb{R}^{K_0\times{K}_1}$, whose columns contain a set of $K_1$ reference positions with dimension $K_0$.
If the columns of matrix $S$ constitute a convex hull for any data point $X(t)$  in the dataset, then, any such data point can be expressed exactly as a convex linear combination of the polytope vertices  $S_{:,k}$ \cite{gerber2020low}:

\begin{equation}\label{eq:conv_hull}
  X(t) = S \{\gamma_{(1)}\}_{:,t}.
\end{equation}

If for all $t=1,\dots,T$, $d=1,\dots,K_0$, $\{\gamma_{(0)}\}_{d,t} > 0 $, and a given input loss function $\textbf{dist}(\cdot,\cdot)$ is additive in dimensions (i.e., for any $x$ and $y$ in $\mathbb{R}^D$, $\textbf{dist}(x,y) = \sum_{d=1}^D \textbf{dist}_d (x_d, y_d)$) and strictly convex, then the exact minimizer of
\begin{equation}\label{eq:layer_first}
  \mathbf{E}_{\gamma_{(0)}} \textbf{dist}(X(\cdot), S\gamma_{1}(\cdot))=\sum_{t,d}^{T,K_0} \{ \gamma_{(0)} \}_{d,t}  \textbf{dist}_d\left(\{X(t)\}_d,  \sum_{k=1}^{K_1} S_{d,k}  \{\gamma_{(1)}\}_{k,t} \right)
\end{equation}
is given by the equation \eqref{eq:conv_hull}.
Since $\{\gamma_{(1)}\}_{:,t}$ is a vector of linear combination coefficients, $\sum_{d=1}^{K_0} \{ \gamma_{(1)} \}_{d,t} = 1$ and $\{\gamma_{(1)}\}_{d,t} \geq 0$ - hence, for every $t$, each column $\{\gamma_{(1)}\}_{d,:}$ is a probability measure.

\paragraph{Entropic layers}
For any given $t$, every subsequent layer $n$ of the network takes the probability distribution $\{\gamma_{(n)}\}_{:,t} \in \mathbb{P}^{K_n}$ as an input from the previous layer and maps it to the output probability distribution $\{\gamma_{(n+1)}\}_{:,t} \in \mathbb{P}^{K_{n+1}}$, deploying the exact law of total probability:
\begin{equation}\label{eq:exact_law}
  \{ \gamma_{(n+1)}\}_{:,t} = \theta^{(n)} \{ \gamma_{(n)}\}_{:,t},
\end{equation}
where $\theta^{(n)} \in \mathbb{R}^{K_{n+1}, K_{n}}$ is a column stochastic matrix of conditional probabilities, providing an exact relation between the random variables described by the distributions $\gamma_{(n)}$ and $\gamma_{(n+1)}$ \cite{schutte2001transfer, schutte2013metastability}. Please note, that here and in the following, the upper index with brackets $\theta^{(n)}$ refers to the index of the layer, while without the round brackets $\theta^\textrm{it}$ it refers to the iteration number in the numerical algorithm.

Then, the exact minimiser of the expected cross-entropy functional
\begin{equation}\label{eq:layer_second}
  \frac{1}{T} \sum_{t=1}^T \mathbf{CE}\left[  \{ \gamma_{(n+1)}\}_{:,t},\theta^{(n)} \{ \gamma_{(n)}\}_{:,t}  \right] =
  - \frac{1}{T} \sum_{t,k=1}^{T,K_{n+1}} \left\{\gamma_{(n+1)}\right\}_{k,t} \left[\log_{K_{n}} \left(\theta^{(n)}_{k,:}\{\gamma_{(n)}\}_{:,t}\right)\right].
\end{equation}

is given by equations \eqref{eq:exact_law}, for all $t=1,\dots,T$, since, for any $x,y \in \mathbb{P}^D$ with non-empty support, $\mathbf{CE} \left[x,y\right]$ is minimal if and only if $x = y$ \cite{cover1999elements}.
The logarithm is taken with the base that corresponds to the dimension of the probability distribution, to ensure that the possible values of the resulting expected cross-entropy are comparable across layers which may have different support and therefore are independent of the distribution dimensions.

\paragraph{Entropic regression layer}
In a regression setting the target value associated with the $t$-th data point $y(t)$ is a real-valued quantity in  $\mathbb{R}$.
Following the same discretization philosophy applied to the inputs (see Paragraph \ref{par:input_layer}), it is possible to obtain an optimal discretization using a finite set of output centroids $S^y \in \mathbb{R}^{K_{N+1}}$, which can be jointly optimized together with the rest of the variables.
The continuous response $y(t)$ can be constructed as the linear combination of those centroids, $S^y \gamma_{(n)}(t)$, where $\gamma_{(n)}(t)$ is the soft assignment vector for the $t$th observation at the last layer.
The loss function term relative to the regression output minimizes the mean distance between the true target and the discretized reconstruction:
\begin{equation}\label{eq:EON_regression_1}
  \sum_{t=1}^{T} \textbf{dist}(y_t, \{ S^y \gamma \}_t).
\end{equation}

Furthermore, in the case of $\textbf{dist}$ being the squared Euclidean distance, the loss term for the output layer, as described in \eqref{eq:EON_regression_1}, can be combined with that of the input layer:
\begin{equation}
  \sum_{t=1}^T \textbf{dist}(X, S\gamma) + \delta \sum_{t=1}^T \textbf{dist}(y, S^y\gamma)
\end{equation}

by considering the following variables:
\begin{equation}\label{eq:regr_change_variable}
  \{ \tilde{X}_\varepsilon \}_{:,t} :=
  \begin{pmatrix}
    X_{:,t}\\ \delta^2 y_t
  \end{pmatrix},
  \{ \tilde{S}_\varepsilon \}_{:,t} :=
  \begin{pmatrix}
    S_{:,k}\\ \delta^2 S^y_k,
  \end{pmatrix}
\end{equation}
where $\delta$ is a tuneable weight controlling the relative importance of the discretization problem of the output compared to that of the input space.

\paragraph{Non-equilibrium entropy constraints}
Note that, as demonstrated in Lemma \eqref{lemma:entropyminimal} below, a linear programming problem subject exclusively to simplex constraints admits an optimal solution with minimal Shannon entropy.
Consequently, obtaining deviations from the entropy minimal point can be enforced with the addition of a (negative) entropy regularization term on the variable within the formulation of the optimization problem.
By varying the associated regularization coefficient, one can continuously interpolate between the extremes of the feasible set: the entropy minimal solution (for small values of the regularization coefficient) and the entropy maximal solution (for larger values of the regularization coefficient). For instance, incorporating entropy regularization on the assignments $\gamma$ yields a continuum of admissible solutions, dependent on the coefficient: from deterministic affiliations (k-means optimal), to uniform discrete distributions, as well as any intermediate distribution.

\begin{lemma}[Entropy-minimal solution of a linear programming problem on the simplex]
  \label{lemma:entropyminimal}
  For any $b \in \mathbb{R}^K$, the solution set of the optimization problem
  \begin{equation}
    \label{eq:linearsimplex}
    \gamma^{*} \in \arg \min\limits_{\gamma \in \mathbb{P}^K} b^\mathsf{T} \gamma
  \end{equation}
  always contains a solution with minimal entropy, i.e., $\mathbf{H}(\gamma^{*}) = 0$.
\end{lemma}

\begin{proof}
  It is easy to verify that the solution to the optimization problem \eqref{eq:linearsimplex} is given by
  \begin{equation}
    \label{eq:linearsimplexsolution}
    \lbrace \gamma^{*} \rbrace_k =
    \left\lbrace
    \begin{array}{ll}
      1 & \text{if } k = \arg\limits_{\hat{k}} \min \lbrace b \rbrace_{\hat{k}} \\
      0 & \text{otherwise},
    \end{array}
    \right.
  \end{equation}
  that is, a solution is constructed by setting the component corresponding to the index of the minimum entry in $b$ to one, while all remaining components are set to zero, in order to satisfy the simplex constraints. In this case, the value of the objective function is minimal among all feasible values. This binary solution has minimal entropy.

  However, if the minimum of $b$ is not unique, i.e., the minimum of
  $b$ is attained at multiple indices, then the set of solutions becomes significantly richer. More precisely, let $\widehat{\mathcal{K}}_{\min}$ denote the set of indices corresponding to the minimum value of $b$, i.e., for all $\hat{k} \in \widehat{\mathcal{K}}_{\min}$ it holds that
  \begin{displaymath}
    \forall k \in \lbrace 1,\dots,K \rbrace \setminus \widehat{\mathcal{K}}_{\min} :
    \quad
    \lbrace b \rbrace_{k} > \lbrace b \rbrace_{\hat{k}},
  \end{displaymath}
  and let $\gamma^{*1}, \dots, \gamma^{*\vert\widehat{\mathcal{K}}_{\min}\vert}$ be the corresponding binary solutions of the form \eqref{eq:linearsimplexsolution}.
  Then, the full solution set of \eqref{eq:linearsimplex} is given by the convex hull of the binary vectors $\gamma^{*1}, \dots, \gamma^{*\vert\widehat{\mathcal{K}}_{\min}\vert}$, i.e., any solution $\gamma^{*}$ can be written as a convex combination of these binary solutions:
  \begin{displaymath}
    \gamma^{*} = \sum\limits_{\hat{k}=1}^{\vert\widehat{\mathcal{K}}_{\min}\vert} \alpha_{\hat{k}} \gamma^{*\hat{k}},
    \quad
    \alpha \in \mathbb{P}^{\vert\widehat{\mathcal{K}}_{\min}\vert}.
  \end{displaymath}
  Nevertheless, even in this case, the solution set includes binary vectors with minimal entropy.

\end{proof}

Therefore, probability distributions $\gamma_{(n)}(t)$ of each layer $(n)$ are subject to the layer-specific expected entropy constraints, which may differ from layer to layer:
\small
\begin{eqnarray}
  \label{eq:EON_H_con}
  \mathbf{H}(\gamma_{(0)})&=&\sum_{t,d=1}^{T,K_0}\left\{\gamma_{(0)}\right\}_{d,t}\log_{K_0T}(\left\{\gamma_{(0)}\right\}_{d,t})\equiv \mathbf{H}_0,  \nonumber\\
  \mathbf{H}(\gamma_{(n)})&=&\frac{1}{T} \sum_{t=1}^{T} \{\gamma_{(n)}\}_{:,t}^\mathsf{T}\log_{K_n}(\{\gamma_{(n)}\}_{:,t})\equiv \mathbf{H}_n, \quad \text{ for all  } n=1,\dots,N+1.
\end{eqnarray}
\normalsize
Also in this case, the logarithm is taken with the base that corresponds to the dimension of the probability distribution, to ensure that the possible values of the resulting normalized entropy $\mathbf{H}$ are confined to the closed interval $\left[0,1\right]$, with the deterministic distributions (i.e., taking only 1/0  values) attaining minimal values $\mathbf{H}=0$ and uniform distributions (having all equal probabilities) attaining maximum value $\mathbf{H}=1$, independent of the dimension $K_n$ of the specific distribution.

\paragraph{Formulation of the classification learning problem}
Combining together the loss functions for the input layer \eqref{eq:layer_first}, the entropic layers \eqref{eq:layer_second} (multiplied by their respective tuneable weights, collected in the vector $\delta=\left(\delta_1,\dots,\delta_N\right)$) together with the entropy constraints \eqref{eq:EON_H_con} (multiplied by their respective regularization parameters, collected in the vector $\epsilon=\left(\epsilon_1,\dots,\epsilon_N\right)$), we obtain the following functional:

\small
\begin{align}\label{eq:EON_fun_nojensen}
  \tilde{L}_{\textbf{EON}}(\Gamma,S,\theta) =&
  \sum\limits_{t=1}^{T}
  \left(
    \underbrace{
      \sum_{d}^{K_0} \{ \gamma_{(0)} \}_{d,t}  \textbf{dist}_d\left(\{X(t)\}_d,  \sum_{k=1}^{K_1} S_{d,k}  \{\gamma_{(1)}\}_{k,t} \right)
    }_{\textrm{input layer}}
    -\underbrace{
      \sum\limits_{n,k=1}^{N,K_{n+1}}
      \frac{\delta_n}{T}
      \left\{\gamma_{(n+1)}\right\}_{k,t} \left[\log_{K_{n}} \left(\theta^{(n)}_{k,:}\{\gamma_{(n)}\}_{:,t}\right)\right]
    }_{\textrm{entropic layers}}
  \right) \\
  & +\underbrace{
    \sum\limits_{n=0}^{N+1}
    \epsilon_n\mathbf{H}(\gamma_{(n)}),
  }_{\textrm{entropy regularization}}\nonumber
\end{align}
\normalsize
where $\Gamma$ represents the collection of $\gamma_{(n)}$ for all $n$, and $\theta$ refers to the collection of $\theta^{(n)}$ for all $n$.

Next, getting use of the convexity of $\textbf{dist}$ and the concavity of the logarithm, we can apply Jensen's inequality to obtain the following constrained optimization problem for learning parameters $[\hat{\Gamma},\hat{\theta},\hat{S}]$, for Jensen's upper bound $L_{\textbf{EON}}(\Gamma,S,\theta) \geq \tilde{L}_{\textbf{EON}}(\Gamma,S,\theta)$:

\small
\begin{eqnarray}
  [\hat{\Gamma},\hat{\theta},\hat{S}]
  &=& \arg \min\limits_{
    \substack{
      \Gamma,\theta,S
    }
  }L_{\textbf{EON}}(\Gamma,S,\theta), \label{eq:EON_opt}\\
  \nonumber
  L_{\textbf{EON}}(\Gamma,S,\theta) &=&
  \sum\limits_{t=1}^{T}
  \left(
    \underbrace{
      \sum_{d,k=1}^{K_0,K_1} \{ \gamma_{(0)} \}_{d,t} \{\gamma_{(1)}\}_{k,t}  \textbf{dist}_d\left(\{X(t)\}_d,  S_{d,k}   \right)
    }_{\textrm{input layer}}
    -\underbrace{
      \sum\limits_{n,k=1}^{N,K_{n+1}}
      \frac{\delta_n}{T}
      \left\{\gamma_{(n+1)}\right\}_{k,t} \log_{K_{n}} \left(\theta^{(n)}_{k,:}\right)\{\gamma_{(n)}\}_{:,t}
    }_{\textrm{entropic layers}}
  \right) \\
  &&+\underbrace{
    \sum\limits_{n=0}^{N+1}
    \epsilon_n\mathbf{H}(\gamma_{(n)})
  }_{\textrm{entropy regularization}},\label{eq:EON_fun}\\
  \label{eq:EON_fun_con1}
  \text{such  that } &&\theta^{(n)}_{:,k}\text{ and } \{\gamma_{(n)}\}_{:,t} \text{ for all $k,n$ $t$ are constrained to be probability distributions}. \\
  \label{eq:EON_fun_con2}
  \text{and } &&\{ \gamma_{(N+1)} \}_{:,t}=\pi(t) \text{, for all $t$ (in the EON training phase)},
\end{eqnarray}
\normalsize
where $\pi(1),\dots,\pi(T)$ is a set of the known $K_{N+1}$-dimensional label probability distributions, that correspond to the features $X(1),\dots,X(T)$ and are used in the training phase only.

\paragraph{Formulation of the regression learning problem}
Problem \eqref{eq:EON_opt} can be extended to regression learning tasks straightforwardly by replacing \eqref{eq:EON_fun} with the following functional, subject to \eqref{eq:EON_fun_con1}:

\small
\begin{align}
  \tilde{L}_{\textbf{EON}}(\Gamma,S,\theta, S^y) =&
  \sum\limits_{t=1}^{T}
  \left(
    \underbrace{
      \sum_{d}^{K_0} \{ \gamma_{(0)} \}_{d,t}  \textbf{dist}_d\left(\{X(t)\}_d,  \sum_{k=1}^{K_1} S_{d,k}  \{\gamma_{(1)}\}_{k,t} \right)
    }_{\textrm{input layer}} \right.\nonumber\\
    & -\left. \underbrace{
      \sum\limits_{n,k=1}^{N-1,K_{n}}
      \frac{\delta_n}{T}
      \left\{\gamma_{(n+1)}\right\}_{k,t} \left[\log_{K_{n}} \left(\theta^{(n)}_{k,:}\{\gamma_{(n)}\}_{:,t}\right)\right]
    }_{\textrm{entropic layers}} \right. \label{eq:EON_fun_nojensen_reg}\\
    &\left.
    +\underbrace{\frac{\delta_N}{T} \textbf{dist} \left(y(t), \sum_{k=1}^{K_{N}}  S^y_k \{ \gamma_{(N)} \}_{k,t} \right)}_{\textrm{regression}}
  \right) +\underbrace{
    \sum\limits_{n=0}^{N}
    \epsilon_n\mathbf{H}(\gamma_{(n)}),
  }_{\textrm{entropy regularization}}\nonumber
\end{align}
\normalsize
where $y(1),\dots,y(T)$ is a set of the known $K_{N+1}$-dimensional real valued targets used in training. Note how, applying the change of variables introduced in Eq. \eqref{eq:regr_change_variable} makes this functional equivalent in form to \eqref{eq:EON_fun_nojensen}.
In a similar way, making use of the convexity of $\textbf{dist}$ and concavity of the logarithm, Jensen's inequality can be applied to the regression problem, obtaining the following Jensen's upper bound  $L_{\textbf{EON}}(\Gamma,S,\theta, S^y) \geq \tilde{L}_{\textbf{EON}}(\Gamma,S,\theta, S^y)$:

\small
\begin{align}
  [\hat{\Gamma},\hat{\theta},\hat{S}, \hat{S^y}]
  &= \arg \min\limits_{
    \substack{
      \Gamma,\theta,S, S^y
    }
  }L_{\textbf{EON}}(\Gamma,S,\theta, S^y), \label{eq:EON_opt_reg}\\
  L_{\textbf{EON}}(\Gamma,S,\theta, S^y) &=
  \sum\limits_{t=1}^{T}
  \left(
    \underbrace{
      \sum_{d,k=1}^{K_0,K_1} \{ \gamma_{(0)} \}_{d,t} \{\gamma_{(1)}\}_{k,t}  \textbf{dist}_d\left(\{X(t)\}_d,  S_{d,k}   \right)
    }_{\textrm{input layer}}
    -\underbrace{
      \sum\limits_{n,k=1}^{N-1,K_{n+1}}
      \frac{\delta_n}{T}
      \left\{\gamma_{(n+1)}\right\}_{k,t} \log_{K_{n}} \left(\theta^{(n)}_{k,:}\right)\{\gamma_{(n)}\}_{:,t}
    }_{\textrm{entropic layers}}
    \right.\nonumber\\
    &\left.
    +\underbrace{\frac{\delta_N}{T}  \sum_{k=1}^{K_{N}} \{  \gamma_{(N)} \}_{k,t} \textbf{dist} \left( y(t),  S^y_k  \right)}_{\textrm{regression}}
  \right) +\underbrace{
    \sum\limits_{n=0}^{N}
    \epsilon_n\mathbf{H}(\gamma_{(n)})
  }_{\textrm{entropy regularization}},\label{eq:EON_fun_regr_jensen}\\
  \label{eq:EON_fun_con1_regr}
  \text{such  that } &\theta^{(n)}_{:,k}\text{ and } \{\gamma_{(n)}\}_{:,t} \text{ for all $k,n$ $t$ are constrained to be probability distributions}.
\end{align}
\normalsize

We will refer to this optimization problem formulation as an Entropy-Optimal feed-forward Network (EON).
As will be shown in the following sections, the Jensen approximation $L_\textbf{EON}$ is essential in order to obtain a solution of this problem that can be implemented as a sequence of straightforward and cheap analytically-computable steps - and without backpropagation and numerical methods based on gradient descent that rely on tuning multiple hyperparameters.

Please note that the obtained expressions in the entropic layers are bilinear forms with respect to the probability measures $\gamma_{(n)}(t)$ and $\gamma_{(n+1)}(t)$ - and not the quadratic forms as in the canonic BM formulation \eqref{eqn:boltzmann}. Also this issue will play an important role below, when we prove the mathematical properties of the problem.
Another significant difference to the equation \eqref{eqn:boltzmann} is that the elements of the matrix defining the bilinear form in this expression are given as logarithms of probability matrices $\theta^{(n)}$ - and, hence, should satisfy additional equality and inequality constraints \eqref{eq:EON_fun_con1} that are not present in the original formulation (\ref{eqn:boltzmann}).
Moreover, tuning the hyperparameters $\delta$ and $\epsilon$ allows achieving different entropy levels at different layers - and enables functioning beyond the equilibrium assumption of the canonic BM.
Please also note that the classic BM formulation \eqref{eqn:boltzmann} does not include a term to handle classification tasks.

\begin{figure}[ht]
  \centering
  \includegraphics[width=0.9\textwidth]{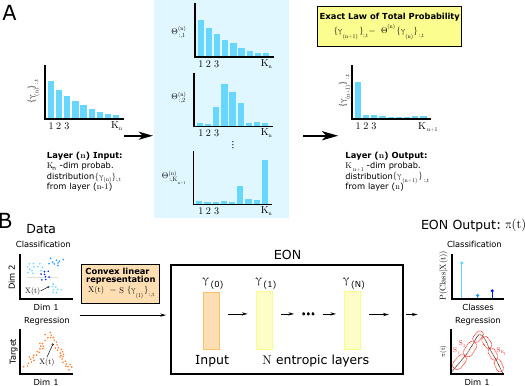}
  \caption{Illustration of the structure of the proposed EON model. (A) Representation of the operation of an entropic layer.
  (B) Given a dataset consisting of $T$ data points $X(t)$, the EON model involves using an input layer and a number of entropic layers. The result is for classification a probability distribution over the possible labels for each data point and for regression a continuous variable.}\label{fig1}
\end{figure}

Another key observation from (\ref{eq:EON_fun}-\ref{eq:EON_fun_con2}), is that for any fixed set of  hyperparameters $\left(K_0,\dots,K_{N+1},\epsilon,\delta\right)$,  (\ref{eq:EON_fun}-\ref{eq:EON_fun_con2}) can be solved analytically for each individual variable $\left(\gamma_{(0)},\dots,\gamma_{(K_{N+1})},\theta,S\right)$ - when leaving the remaining  $(K_{N+1}+3)$ variables fixed; the solutions are described in the following sections, while the coordinate descent algorithm that minimizes \ref{eq:EON_fun}-\ref{eq:EON_fun_con2} is presented as pseudocode in Algorithm \ref{alg:subspace}.
Lemma \ref{lemma:generalsubspace} establishes the fundamental convergence of a general coordinate descent algorithm and can be directly applied to demonstrate the minimization property of Algorithm \ref{alg:subspace}.

\begin{algorithm}
  \caption{Coordinate descent algorithm (please note: upper index refers to iteration number)}\label{alg:subspace}
  \KwData{$X \in \mathbb{R}^{K_0,T}, \pi \in \mathbb{P}_L^{K_{N+1},T}$}
  \KwResult{$\hat{\Gamma} \approx \Gamma^{\iit},\hat{\theta} \approx \theta^{\iit},\hat{S} \approx S^{\iit}$}
  \vspace{2mm}
  $\textrm{choose}~ \Gamma^0 \in \Omega_{\Gamma}, S^0 \in \Omega_{S}, \theta^0 \in \Omega_{\theta}$ \Comment*[r]{initial approximation}
  $\iit \gets 0$ \Comment*[r]{iteration counter}
  \vspace{2mm}
  \While{the stopping criteria is not satisfied}{
    \vspace{2mm}
    $\Gamma^{\iit+1} \gets \arg \min\limits_{\Gamma \in \Omega_{\Gamma}} L_{\textbf{EON}}(\Gamma, S^{\iit},\theta^{\iit})$ \Comment*[r]{solve $\Gamma$-problem}
    $S^{\iit+1} \gets \arg \min\limits_{S \in \Omega_{S}} L_{\textbf{EON}}(\Gamma^{\iit+1},S, \theta^{\iit})$ \Comment*[r]{solve $S$-problem}
    $\theta^{\iit+1} \gets \arg \min\limits_{\theta \in \Omega_{\theta}} L_{\textbf{EON}}(\Gamma^{\iit+1},S^{\iit+1}, \theta)$ \Comment*[r]{solve $\theta$-problem}
    $\iit \gets \iit + 1$\;
  }
  \vspace{2mm}
\end{algorithm}

Briefly, the algorithm starts with an initial guess for the variables $\left(\theta,S\right)$, and keeps iterating using exact analytical solutions until the change of the value $L_{\textbf{EON}}$ in one iteration is less then a predefined tolerance threshold. In every iteration, the EON algorithm considers every optimization variable, one by one, and moves to the optimum of the subproblem with respect to this variable. Hence, the training procedure for EON models is not subject to the problem of under- or over-shooting the local loss optimum (as is instead the case for the common GD and SGD methods)  - and does not rely on tuning the respective gradient-descent-related numerical algorithm parameters like learning rate, momentum parameters, batch size, etc.

\begin{lemma}[General coordinate descent algorithm monotonicity]
  \label{lemma:generalsubspace}
  Let us consider a general optimization problem
  \begin{equation}
    [\hat{x}_{(1)},\dots,\hat{x}_{(M)}] = \arg \min f(x_{(1)},\dots,x_{(M)}) ~~ \textrm{subject to} ~~ \forall m: x_{(m)} \in \Omega_m.
  \end{equation}
  The sequence generated by coordinate descent algorithm (for iterations $\iit = 0,1,2,\dots$ and any initial $x^{\mathrm{0}} := [x_{(1)}^{\mathrm{0}},\dots,x_{(M)}^{\mathrm{0}}] \in \Omega_1 \times \dots \times \Omega_M$)
  \begin{equation}
    \label{eq:generalsubspace}
    \textrm{for} ~~~ m = 1,\dots,M ~~~\textrm{solve}~~~
    x_{(m)}^{\iit+1} = \arg \min\limits_{x \in \Omega_m}
    f(x_{(1)}^{\iit+1},\dots,x_{(m-1)}^{\iit+1},x,x_{(m+1)}^{\iit},\dots,x_{(M)}^{\iit})
  \end{equation}
  is providing the monotone non-increase of objective function $f$, i.e.,
  \begin{equation}
    \label{eq:generalsubspace_decrease}
    f(x^{\iit + 1}) \leq f(x^{\iit}).
  \end{equation}
\end{lemma}

\begin{proof}
  From the zero-order optimality condition of minimization problem \eqref{eq:generalsubspace}, we have
  \begin{equation}
    \label{eq:generalsubspace_aux1}
    \forall x\in \Omega_m:
    f(x_{(1)}^{\iit+1},\dots,x_{(m-1)}^{\iit+1},x,x_{(m+1)}^{\iit},\dots,x_{(M)}^{\iit})
    \geq
    f(x_{(1)}^{\iit+1},\dots,x_{(m-1)}^{\iit+1},x_{(m)}^{\iit+1},x_{(m+1)}^{\iit},\dots,x_{(M)}^{\iit})
    .
  \end{equation}
  In this inequality, we choose $x := x_{(m)}^{\iit} \in \Omega_m$ to get
  \begin{equation}
    \label{eq:generalsubspace_aux2}
    f(x_{(1)}^{\iit+1},\dots,x_{(m-1)}^{\iit+1},x_{(m)}^{\iit},x_{(m+1)}^{\iit},\dots,x_{(M)}^{\iit}) \geq
    f(x_{(1)}^{\iit+1},\dots,x_{(m-1)}^{\iit+1},x_{(m)}^{\iit+1},x_{(m+1)}^{\iit},\dots,x_{(M)}^{\iit}),
  \end{equation}
  which can be recursively used to prove inequality \eqref{eq:generalsubspace_decrease}
  \begin{displaymath}
    \begin{array}{rcl}
      f(x^{\iit}) & = & f(x_{(1)}^{\iit},x_{(2)}^{\iit},x_{(3)}^{\iit}, \dots, x_{(M-1)}^{\iit},x_{(M)}^{\iit}) \\[2mm]
      & \geq & f(x_{(1)}^{\iit+1},x_{(2)}^{\iit},x_{(3)}^{\iit}, \dots, x_{(M-1)}^{\iit},x_{(M)}^{\iit}) \\[2mm]
      & \geq & f(x_{(1)}^{\iit+1},x_{(2)}^{\iit+1},x_{(3)}^{\iit}, \dots, x_{(M-1)}^{\iit},x_{(M)}^{\iit}) \\[2mm]
      & \vdots & \\[2mm]
      & \geq & f(x_{(1)}^{\iit+1},x_{(2)}^{\iit+1},x_{(3)}^{\iit+1}, \dots, x_{(M-1)}^{\iit+1},x_{(M)}^{\iit}) \\[2mm]
      & \geq & f(x_{(1)}^{\iit+1},x_{(2)}^{\iit+1},x_{(3)}^{\iit+1}, \dots, x_{(M-1)}^{\iit+1},x_{(M)}^{\iit+1}) = f(x^{\iit + 1}).
    \end{array}
  \end{displaymath}
\end{proof}

\subsection{EON with the squared Euclidean distance}
In the following section, we demonstrate that when using squared Euclidean distance in the input layer to measure the discrepancy between feature components $X(t)$ and feature patterns in the columns of $S$ (and, for regression problems, also in the output layer to measure the distance between $y(t)$ and $S^y \gamma_{(N)}$), the exact solutions to all subproblems in the above Algorithm 1 can be derived analytically, and the computational complexity can be determined.
In this case, the problem formulation \eqref{eq:EON_opt} can be expanded into
\begin{equation}
  \label{eq:EON_fun_E2}
  \begin{array}{rcl}
    L(\Gamma,\theta,S) & = &
    \sum\limits_{t=1}^{T}
    \left(
      \underbrace{
        \sum\limits_{k_1=1}^{K_1} \lbrace \gamma_{(1)} \rbrace_{k_1,t}
        \sum\limits_{d=1}^{K_0} \lbrace \gamma_{(0)} \rbrace_{d,t} ~ \big( \lbrace X(t) \rbrace_{d} - S_{d,k_1} \big)^2
      }_{\textrm{input layer}}
      \right.
      \\
      & & ~~~~~~~~ -
      \underbrace{
        \sum\limits_{n=1}^{N}
        \delta_n
        \left(
          \sum\limits_{k_n = 1}^{K_n} \sum\limits_{k_{n+1} = 1}^{K_{n+1}}
          \lbrace \gamma_{(n)} \rbrace_{k_n,t}
          \lbrace \gamma_{(n+1)} \rbrace_{k_{n+1},t}
          \log (\theta^{(n)}_{k_n,k_{n+1}})
        \right)
      }_{\textrm{entropic layers}}
      \\
      & &
      \left.
      ~~~~~~~~ +
      \underbrace{
        \sum\limits_{n=0}^{N+1}
        \epsilon_n \left( \sum\limits_{k_n=1}^{K_n} \lbrace \gamma_{(n)} \rbrace_{k_n,t} \log \lbrace \gamma_{(n)} \rbrace_{k_n,t}
        \right)
      }_{\textrm{entropy regularization}}
    \right).
  \end{array}
\end{equation}

For regression learning tasks, the functional \eqref{eq:EON_fun_regr_jensen} can be expressed as:

\begin{align}
  L(\Gamma,S,\theta, S^y) =&
  \sum\limits_{t=1}^{T}
  \left(
    \underbrace{
      \sum_{d,k=1}^{K_0,K_1} \{ \gamma_{(0)} \}_{d,t} \{\gamma_{(1)}\}_{k,t}  \left(\{X(t)\}_d -  S_{d,k}   \right)^2
    }_{\textrm{input layer}} \right. -\left. \underbrace{
      \sum\limits_{n,k=1}^{N-1,K_{n}}
      \frac{\delta_n}{T}
      \left\{\gamma_{(n+1)}\right\}_{k,t} \log_{K_{n}} \left(\theta^{(n)}_{k,:}\right)\{\gamma_{(n)}\}_{:,t}
    }_{\textrm{entropic layers}}
    \right.\label{eq:EON_fun_regr_jensen_euclid}\\
    &+\left.
    \underbrace{\frac{\delta_{N}}{T}  \sum_{k=1}^{K_{N}} \{  \gamma_{(N)} \}_{k,t} \left( y(t) -  S^y_k  \right)^2}_{\textrm{regression}}
  \right) +\underbrace{
    \sum\limits_{n=0}^{N}
    \epsilon_n \left( \sum\limits_{k_n=1}^{K_n} \lbrace \gamma_{(n)} \rbrace_{k_n,t} \log \lbrace \gamma_{(n)} \rbrace_{k_n,t}
        \right).
  }_{\textrm{entropy regularization}}\nonumber
\end{align}

For both classification and the regression, the formulation includes as feasible sets
\begin{equation}
  \label{eq:Omega_E2}
  \Omega_{\Gamma} :=
  \left(~
    \Omega_{\gamma_{(0)}}
    \times
    \mathbb{P}_{\textrm{L}}^{K_1 \times T} \times \dots \times \mathbb{P}_{\textrm{L}}^{K_{N+1} \times T}
  ~\right), ~~~
  \Omega_{\theta} :=
  \left(~
    \mathbb{P}^{K_1 \times K_2}_{\textrm{L}} \times \dots \times \mathbb{P}^{K_{N} \times K_{N+1}}_{\textrm{L}}
  ~\right), ~~~
  \Omega_{S} := \mathbb{R}^{K_0 \times K_1},
\end{equation}
and given positive regularization hyperparameters $\delta \in \mathbb{R}^{N}$ and
$\epsilon \in \mathbb{R}^{N+1}$. For a more formal description of the feasible set $\Omega_{\gamma_{(0)}}$, we defer the explanation to the paragraph \ref{par:sol_gamma} - "Solution of the $\Gamma$ problem" below.

\paragraph{Solution of the $S$ problem}

\begin{lemma}[$S$-problem: analytic solution]
  \label{lemma:Sproblem}
  With given $\Gamma \in \Omega_{\Gamma}$ and $\theta \in \Omega_{\theta}$, the optimization problem \eqref{eq:EON_fun_E2} has analytic solution (for every $\hat{d} = 1,\dots,K_0, ~ \hat{k}_1 = 1,\dots,K_1$)
  \begin{equation}
    \label{eq:EON_solS}
    \hat{S}_{\hat{d},\hat{k}_1} =
    \frac{
      \sum\limits_{t=1}^{T}
      \lbrace \gamma_{(0)} \rbrace_{\hat{d},t}
      \lbrace \gamma_{(1)} \rbrace_{\hat{k}_1,t} \lbrace X(t) \rbrace_{\hat{d}}
    }{
      \sum\limits_{t=1}^{T}
      \lbrace \gamma_{(0)} \rbrace_{\hat{d},t}
      \lbrace \gamma_{(1)} \rbrace_{\hat{k}_1,t}
    }
    .
  \end{equation}
  The evaluation complexity of this solution is $\mathcal{O}(T K_0 K_1)$.
\end{lemma}

\begin{proof}
  With given $\Gamma$ and $\theta$, the optimization problem \eqref{eq:EON_opt}
  can be reduced to unconstrained problem
  \begin{displaymath}
    \hat{S}
    = \arg \min\limits_{
      S \in \mathbb{R}^{K_0 \times K_1}
    } L_{\Gamma,\theta}(S),
  \end{displaymath}
  where the objective function
  \begin{displaymath}
    L_{\Gamma,\theta}(S) =
    \frac{1}{T}
    \sum\limits_{t=1}^{T}
    \sum\limits_{k_1=1}^{K_1} \lbrace \gamma_{(1)} \rbrace_{k_1,t}
    \sum\limits_{d=1}^{K_0} \lbrace \gamma_{(0)} \rbrace_{d,t}
    \left(
      \lbrace X(t) \rbrace_d - S_{d,k_1}
    \right)^2
  \end{displaymath}
  is continuously differentiable. To derive the solution from first-order necessary optimality condition (derivative equal to zero), we derive partial derivatives (for any $\hat{d} = 1,\dots,K_0$ and any $\hat{k}_1 = 1,\dots,K_1$)
  \begin{displaymath}
    \frac{\partial L_{\Gamma,\theta}(S)}{\partial S_{\hat{d},\hat{k}}}
    =
    -
    \frac{2}{T}
    \sum\limits_{t=1}^{T}
    \lbrace \gamma_{(0)} \rbrace_{\hat{d},t}
    \lbrace \gamma_{(1)} \rbrace_{\hat{k}_1,t}
    \left(
      \lbrace X(t) \rbrace_{\hat{d}} - S_{\hat{d},\hat{k}_1}
    \right)
  \end{displaymath}
  and set them equal to zero. We obtain the solution \eqref{eq:EON_solS}. This solution is well-defined as long as
  \begin{displaymath}
    \sum\limits_{t=1}^{T}
    \lbrace \gamma_{(0)} \rbrace_{\hat{d},t}
    \lbrace \gamma_{(1)} \rbrace_{\hat{k}_1,t}
    \neq 0,
  \end{displaymath}
  i.e., the clusters are non-empty. If a cluster is empty, then the corresponding values $\hat{S}_{:,\hat{k}_1}$ do not appear in the optimization problem and their values can be chosen arbitrary.

  The problem is convex, which can be proven using the second derivative. Indeed, for any $\hat{k}_1$, the Hessian matrix $\nabla^2_{S_{:,\hat{k}_1}} L_{\Gamma,\theta}(S)$
  is a diagonal matrix with non-negative diagonal elements and strictly positive as long as clusters are non-empty.
  The estimation of the computational complexity of this step is trivial - $K_0 K_1$ components of the solution $\hat{S} \in \mathbb{R}^{K_0 \times K_1}$ need to be evaluated and, in each evaluation, the sum is computed through $T$.
\end{proof}

\paragraph{Solution of the $S^y$ problem}
The optimisation subproblem for the output centroids $S^y$, which arises only in regression tasks, is structurally very similar to the one for input centroids. Consequently, its solution can be trivially obtained in closed form by applying the same weighted averaging procedure, with the exception that $\gamma_{(0)}$ is omitted:
\begin{equation}
  \label{eq:EON_solSy}
  \hat{S^y}_{\hat{k}_n} =
  \frac{
    \sum\limits_{t=1}^{T}
    \lbrace \gamma_{(n)} \rbrace_{\hat{k}_1,t} y(t)
  }{
    \sum\limits_{t=1}^{T}
    \lbrace \gamma_{(n)} \rbrace_{\hat{k}_1,t}
  }
  .
\end{equation}

The evaluation cost for this step is also trivial, as $K_n$ components need to be evaluated, through a single pass over the training set, resulting in a $\mathcal{O}(TK_n)$ cost complexity.

\paragraph{Solution of the $\theta$ problem}

\begin{lemma}[$\theta$-problem: analytic solution]
  \label{lemma:thetaproblem}
  With given $\Gamma$ and $S$, the optimization problem \eqref{eq:EON_fun_E2} has analytic solution (for every $n = 1,\dots,N, ~ \hat{k}_{n} = 1,\dots,K_n, ~ \hat{k}_{n+1} = 1,\dots,K_{n+1}$)
  \begin{equation}
    \label{eq:EON_soltheta}
    \hat{\theta}^{(n)}_{\hat{k}_n,\hat{k}_{n+1}}
    = \dfrac{
      \sum\limits_{t=1}^{T}
      \lbrace \gamma_{(n)} \rbrace_{\hat{k}_n,t}
      \lbrace \gamma_{(n+1)} \rbrace_{\hat{k}_{n+1},t}
    }
    {
      \sum\limits_{k_{n} = 1}^{K_{n}}
      \sum\limits_{t=1}^{T}
      \lbrace \gamma_{(n)} \rbrace_{k_n,t}
      \lbrace \gamma_{(n+1)} \rbrace_{\hat{k}_{n+1},t}
    }
  \end{equation}
  The evaluation complexity of this solution is $\mathcal{O}(T K_{\textrm{all}})$, where $K_{\textrm{all}}$ is a number of all transitions between layers given by
  \begin{equation}
    \label{eq:EON_solLambda_Kall}
    K_{\textrm{all}} = \sum\limits_{n=1}^N K_{n} K_{n+1}.
  \end{equation}
\end{lemma}

\begin{proof}
  With given $\Gamma$ and $S$, the optimization problem \eqref{eq:EON_opt}
  can be reduced to constrained problem
  \begin{displaymath}
    \hat{\theta}
    = \arg \min\limits_{
      \theta \in \Omega_{\theta}
    } L_{\Gamma,S}(\theta),
  \end{displaymath}
  with objective function
  \begin{displaymath}
    L_{\Gamma,S}(\theta) =
    -
    \frac{1}{T}
    \sum\limits_{t=1}^{T}
    \sum\limits_{n=1}^{N}
    \delta_n
    \left(
      \sum\limits_{k_n = 1}^{K_n} \sum\limits_{k_{n+1} = 1}^{K_{n+1}}
      \lbrace \gamma_{(n)} \rbrace_{k_n,t}
      \lbrace \gamma_{(n+1)} \rbrace_{k_{n+1},t}
      \log(\theta^{(n)}_{k_{n},k_{n+1}})
    \right).
  \end{displaymath}

  The problem is separable in $n$ (because of separability of both of objective function and feasible set), i.e., the transition matrix for each layer can be evaluated individually. For every $n = 1,\dots,N$ we solve the following optimization problem (positive constant in the objective function are omitted as it does not have any impact on the minimizer)
  \begin{displaymath}
    \hat{\theta}^{(n)}
    = \arg \min\limits_{
      \theta^{(n)}
      \in \mathbb{P}^{K_{n} \times K_{n+1}}
    } L^{(n)}_{\Gamma,S}(\theta^{(n)}),
    ~~~
    L^{(n)}_{\Gamma,S}(\theta^{(n)}) :=
    -
    \sum\limits_{t=1}^{T}
    \left(
      \sum\limits_{k_{n} = 1}^{K_{n}} \sum\limits_{k_{n+1} = 1}^{K_{n+1}}
      \lbrace \gamma_{(n)} \rbrace_{k_n,t}
      \lbrace \gamma_{(n+1)} \rbrace_{k_{n+1},t}
      \log(\theta^{(n)}_{k_{n},k_{n+1}})
    \right).
  \end{displaymath}

  In the derivation of the solution, the inequality constraints will be ignored, but for the derived solution, these constraints are satisfied automatically. The corresponding Lagrange function has form
  \begin{displaymath}
    \mathcal{L}
    (\theta^{(n)},\mu) =
    -
    \sum\limits_{t=1}^{T}
    \left(
      \sum\limits_{k_n = 1}^{K_{n}} \sum\limits_{k_{n+1} = 1}^{K_{n+1}}
      \lbrace \gamma_{(n)} \rbrace_{k_n,t}
      \lbrace \gamma_{(n+1)} \rbrace_{k_{n+1},t}
      \log(\theta^{(n)}_{k_{n},k_{n+1}})
    \right)
    +
    \sum\limits_{k_{n+1}=1}^{K_{n+1}} \mu_{k_{n+1}}
    \left(
      \sum\limits_{k_n=1}^{K_{n}} \theta^{(n)}_{k_{n},k_{n+1}} - 1
    \right)
  \end{displaymath}
  where $\mu \in \mathbb{R}^{K_{n+1}}$ are Lagrange multipliers corresponding to equality constraints of the feasible set $\mathbb{P}_L^{K_n \times K_{n+1}}$ (each column sums to one). The Karush-Kuhn-Tucker (KKT) optimality conditions can be derived as
  \begin{displaymath}
    \begin{array}{rclcl}
      \forall \hat{k}_n, \hat{k}_{n+1}: ~~
      \dfrac{\partial \mathcal{L}(\theta^{(n)},\mu)}{\partial
        \theta^{(n)}_{\hat{k}_n,\hat{k}_{n+1}}
      } & = &
      -
      \dfrac{\sum\limits_{t=1}^{T}
        \lbrace \gamma_{(n)} \rbrace_{\hat{k}_n,t}
        \lbrace \gamma_{(n+1)} \rbrace_{\hat{k}_{n+1},t}
      }{\theta^{(n)}_{\hat{k}_n,\hat{k}_{n+1}}} + \mu_{\hat{k}_{n+1}} ~~~ & = & 0, \\[5mm]
      \forall \hat{k}_n: ~~
      \dfrac{\partial \mathcal{L}(\theta^{(n)},\mu)}{\partial
        \mu_{\hat{k}_{n+1}}
      } & = &
      \sum\limits_{k_{n}=1}^{K_{n}} \theta^{(n)}_{k_{n},\hat{k}_{n+1}} - 1 & = & 0.
    \end{array}
  \end{displaymath}
  From the first equation, we get
  \begin{equation}
    \label{eq:theta_der}
    \lbrace \theta^{(n)} \rbrace_{\hat{k}_n,\hat{k}_{n+1}}
    = \dfrac{
      \sum\limits_{t=1}^{T}
      \lbrace \gamma_{(n)} \rbrace_{\hat{k}_n,t}
      \lbrace \gamma_{(n+1)} \rbrace_{\hat{k}_{n+1},t}
    }{\mu_{\hat{k}_{n+1}}},
  \end{equation}
  which after the substitution into the second KKT results to
  \begin{displaymath}
    \sum\limits_{k_{n}=1}^{K_{n}}
    \dfrac{\sum\limits_{t=1}^{T}
      \lbrace \gamma_{(n)} \rbrace_{k_n,t}
      \lbrace \gamma_{(n+1)} \rbrace_{\hat{k}_{n+1},t}
    }{
      \mu_{\hat{k}_{n+1}}
    } - 1 = 0
    ~~~~
    \Rightarrow
    ~~~~
    \mu_{\hat{k}_{n+1}} =
    \sum\limits_{k_{n}=1}^{K_{n}}
    \sum\limits_{t=1}^{T}
    \lbrace \gamma_{(n)} \rbrace_{k_n,t}
    \lbrace \gamma_{(n+1)} \rbrace_{\hat{k}_{n+1},t}
  \end{displaymath}
  and this can be substituted back to \eqref{eq:theta_der} to obtain \eqref{eq:EON_soltheta}. Obviously, this solution satisfies inequality constraints of the feasible set $\mathbb{P}^{K_{n+1} \times K_n}$.

  The evaluation of the solution \eqref{eq:EON_soltheta} consist in the computation of two sums through $T$ for every component of $\hat{\theta}^{(n)} \in \mathbb{P}^{K_{n} \times K_{n+1}} \subset \mathbb{R}^{K_{n} \times K_{n+1}}$. The sum of all these computations gives the estimation of evaluation complexity.
\end{proof}

\paragraph{Solution of the $\Gamma$ problem}\label{par:sol_gamma}
In the case of the $\Gamma$-problem, the solution depends on the choice of $\gamma_{(0)}$. The feasible set of variable $\gamma_{(0)} \in \Omega_{\gamma_{(0)}}$ is given by the choice of different possible settings:
\begin{itemize}
  \item[Case 1:] All data instances and dimensions are equally important. We set
    \begin{equation}
      \label{eq:gamma0_case1}
      \forall d = 1,\dots,K_0 ~
      \forall t = 1,\dots,T: ~
      \lbrace \gamma_{(0)} \rbrace_{d,t} := \frac{1}{TK_0}
    \end{equation}
    and consequently, $\gamma_{(0)}$ is not unknown.
  \item[Case 2:] We consider a pure feature selection problem. In this case, we are searching for vector $w \in \mathbb{P}^{K_0}$ such that
    \begin{equation}
      \label{eq:gamma0_case2}
      \gamma_{(0)} := \frac{1}{T} w \onebb_T^{\mathsf{T}}, ~~ \textrm{i.e.,} ~~
      \forall d = 1,\dots,K_0 ~
      \forall t = 1,\dots,T: ~
      \lbrace \gamma_{(0)} \rbrace_{d,t} = \frac{1}{T} \lbrace w \rbrace_d.
    \end{equation}
    Components of the unknown stochastic vector $w$ defines the importance of individual data dimensions. The corresponding feasible set is given by
    \begin{equation}
      \Omega_{\gamma_{(0)}} := \mathbb{P}^{K_0}.
    \end{equation}
  \item[Case 3:] We the consider feature importance vector $w \in \mathbb{P}^{K_0}$ and instance importance vector $s \in \mathbb{P}^{T}$. Then
    \begin{equation}
      \label{eq:gamma0_case3}
      \gamma_{(0)} := w s^{\mathsf{T}}, ~~ \textrm{i.e.,} ~~
      \forall d = 1,\dots,K_0 ~
      \forall t = 1,\dots,T: ~
      \lbrace \gamma_{(0)} \rbrace_{d,t} = \lbrace w \rbrace_d \lbrace s \rbrace_t.
    \end{equation}
    The goal of the learning phase is to determine the unknown vectors $w$ and $s$.
    The feasible set is given by
    \begin{equation}
      \Omega_{\gamma_{(0)}} := \mathbb{P}^{K_0} \times \mathbb{P}^{T}.
    \end{equation}
  \item[Case 4:] We assume that the given data have heterogeneous feature importance, and therefore $\gamma_{0}$ is an unknown arbitrary probability matrix.
\end{itemize}

One important consequence of the introduction of $\gamma_{(0)}$ is that it allows to directly obtain the domain definition of the training features $\Omega$. Indeed, $\gamma_{(0)}$ can be considered to be a prior probability distribution function, as well as a function that maps from the domain of the input data (e.g., the unit hypercube for min-max scaled data) to the probability of being in $\Omega$, i.e., $ \gamma_{(0)} : [0,1]^{K_0} \mapsto [0,1]$.
The four cases mentioned above describe how to select a prior based on the contribution of each data point. Alternatively, this value can be estimated from the data itself.
This fact opens the possibility of performing learning in the presence of data displaying missing features. In this case, provided that the setting matches case 4, or that the entries corresponding to missing features are excluded from the optimization problem, the corresponding value of $\gamma_{(0)}$ can simply be set to be equal to $0$ and kept fixed throughout the optimization, which will lead to finding an optimal solution of the problem, and will effectively lead to the exclusion of the features for which a measurement is absent from the optimization procedure.

In all of the cases presented above, the solution exists, as shown in the following lemma.

\begin{lemma}[$\Gamma$-problem: solvability]
  \label{lemma:Gamma_solvability}
  The solution of the minimization problem \eqref{eq:EON_opt} with fixed variables $S$ and $\theta$ always exists.
\end{lemma}

\begin{proof}
  Adopting the convenient notation
  \begin{displaymath}
    0 \log 0 = \lim_{x \rightarrow 0+} \left( x \log x \right) = 0,
  \end{displaymath}
  for any fixed $S \in \Omega_S$ and $\theta \in \Omega_\theta$, the objective function is continuous on a given compact set $\Omega_\Gamma$. The existence of minimizer is given by Weierstrass extreme value theorem.
\end{proof}

Furthermore, we will make use of the following lemma, which was originally introduced in \cite{vecchiESPAScalableEntropyOptimal2022a}, reported below for completeness.

\begin{lemma}[LP with entropy: analytical solution]
  \label{lemma:LP_with_entropy}
  Let $b \in \mathbb{R}^K$ with problem dimension $K \in \mathbb{N}$, and $\epsilon > 0$ is a given regularization parameter. Then the solution of optimization problem
  \begin{equation}
    \label{eq:LPentropy}
    \hat{w} = \arg \min\limits_{w \in \mathbb{P}^K} \langle w, b \rangle + \epsilon
    \langle w, \log w \rangle
  \end{equation}
  has unique analytic solution
  \begin{equation}
    \label{eq:LPentropy_solution}
    \hat{w} = \softmax \left( -\epsilon^{-1} b \right).
  \end{equation}
  The evaluation complexity of this solution is $\mathcal{O}(K)$.
\end{lemma}

\begin{proof}
  See \cite{vecchiESPAScalableEntropyOptimal2022a}, Theorem 1. For the evaluation of the softmax\footnote{
  In our implementation, we use the softmax function provided by standard BLAS libraries, which incorporates the shifting invariance of softmax for improved numerical stability. Specifically, the maximum value in the input vector is subtracted from all elements before computing the exponential. Despite this adjustment, the overall computational complexity remains linear.}
  function in \eqref{eq:LPentropy_solution}, one has to compute the value of exponential function in the nominator of the formula for individual components
  and then compute the sum of nominators to provide the denominator, i. e.,
  \begin{displaymath}
    \textrm{for} ~ k=1,\dots,K:
    \lbrace w \rbrace_k := \exp (-\epsilon^{-1}\lbrace b \rbrace_k), ~~
    w_{\mathrm{sum}} := \sum\limits_{k=1}^K \lbrace w \rbrace_k, ~~
    \lbrace \hat{w} \rbrace_k := \frac{\lbrace w \rbrace_k}{w_{\mathrm{sum}}}.
  \end{displaymath}
  This computation linearly depends on the dimension of the problem.
\end{proof}

\begin{lemma}[Auxiliary]
  \label{lemma:absest}
  (auxiliary)
  If $a, b, c, d > 0$ then
  \begin{equation}
    \label{eq:absest}
    \left\vert
    \frac{a}{b} - \frac{c}{d}
    \right\vert
    \leq
    \frac{\vert a - c \vert}{b}
    +
    \frac{c}{bd}
    \vert b - d
    \vert
    .
  \end{equation}
\end{lemma}

\begin{proof}
  Using a smart zero, we can write and further modify
  \begin{displaymath}
    \frac{a}{b}
    -
    \frac{c}{d}
    =
    \left(
      \frac{a}{b}
      -
      \frac{c}{b}
    \right)
    +
    \left(
      \frac{c}{b}
      -
      \frac{c}{d}
    \right)
    =
    \frac{a - c}{b}
    +
    c
    \left(
      \frac{1}{b}
      -
      \frac{1}{d}
    \right).
  \end{displaymath}
  Now using a triangle inequality, we can estimate
  \begin{displaymath}
    \left\vert
    \frac{a}{b}
    -
    \frac{c}{d}
    \right\vert
    \leq
    \left\vert
    \frac{a - c}{b}
    \right\vert
    +
    \left\vert
    c
    \left(
      \frac{1}{b}
      -
      \frac{1}{d}
    \right)
    \right\vert
    =
    \frac{\vert a - c \vert}{b}
    +
    c
    \left\vert
    \frac{1}{b}
    -
    \frac{1}{d}
    \right\vert.
  \end{displaymath}
  Additionally, the second term can be written in form
  \begin{displaymath}
    \left\vert
    \frac{1}{b}
    -
    \frac{1}{d}
    \right\vert
    =
    \left\vert
    \frac{d - b}{db}
    \right\vert
    =
    \frac{\vert b - d \vert}{db},
  \end{displaymath}
  which gives the final form of upper bound.
\end{proof}

\begin{lemma}[about exponential function]
  \hspace*{0cm}\\
  \begin{itemize}
    \item
      For any $x,y \in [x_{\min},x_{\max}]^K \subset \mathbb{R}^K$ it holds
      \begin{equation}
        \label{eq:expbound}
        e^{x_{\min}} \leq e^{\lbrace x \rbrace_i} \leq e^{x_{\max}},
      \end{equation}

      \begin{equation}
        \label{eq:expsumbound_bound}
        Ke^{x_{\min}} \leq \sum\limits_{j=1}^K e^{\lbrace x \rbrace_j} \leq Ke^{x_{\max}},
      \end{equation}

      \begin{equation}
        \label{eq:expL}
        \vert e^{\lbrace x \rbrace_i} - e^{\lbrace y \rbrace_i} \vert \leq e^{x_{\max}} \vert \lbrace x \rbrace_i - \lbrace y \rbrace_i \vert,
      \end{equation}

      \begin{equation}
        \label{eq:expsumL}
        \left\vert
        \sum\limits_{j=1}^K e^{\lbrace x \rbrace_j}
        -
        \sum\limits_{j=1}^K e^{\lbrace y \rbrace_j}
        \right\vert
        \leq
        e^{x_{\max}}
        \Vert x - y \Vert_1.
      \end{equation}

    \item
      For any $x,y \in \mathbb{P}^K$ it holds
      \begin{equation}
        \label{eq:expbound2}
        \lbrace x \rbrace_i + 1 \leq e^{\lbrace x \rbrace_i} \leq (e-1) \lbrace x \rbrace_i + 1,
      \end{equation}

      \begin{equation}
        \label{eq:expsumbound}
        K + 1 \leq \sum\limits_{j=1}^K e^{\lbrace x \rbrace_j} \leq K + (e - 1).
      \end{equation}
  \end{itemize}

\end{lemma}

\begin{proof}
  \hspace{0cm}\\
  \begin{itemize}
    \item The inequality \eqref{eq:expbound} is trivial and results from the monotonicity of exponential function. The inequality \eqref{eq:expsumbound_bound} can be derived by application of \eqref{eq:expbound} to every term in the sum.
      The equation \eqref{eq:expL} is a Lipschitz inequality of exponential function $\varphi(x) := e^x$ on the interval $[x_{\min},x_{\max}]$. The Lipschitz constant $L_{e^x}$ can be estimated by the maximum value of the derivative of $\varphi$ on the given interval, i.e.,
      \begin{displaymath}
        L_{e^x}
        \leq
        \max\limits_{x \in [x_{\min},x_{\max}]} \vert \varphi'(x) \vert
        =
        \max\limits_{x \in [x_{\min},x_{\max}]} \vert e^x \vert
        = e^{x_{\max}}.
      \end{displaymath}
      The inequality \eqref{eq:expsumL} can be obtained using triangle inequality and \eqref{eq:expL}
      \begin{displaymath}
        \left\vert
        \sum\limits_{j=1}^K
        \left(
          e^{\lbrace x \rbrace_j}
          -
          e^{\lbrace y \rbrace_j}
        \right)
        \right\vert
        \leq
        \sum\limits_{j=1}^K
        \left\vert
        e^{\lbrace x \rbrace_j}
        -
        e^{\lbrace y \rbrace_j}
        \right\vert
        \leq
        \sum\limits_{j=1}^K
        e^{x_{\max}}
        \left\vert
        \lbrace x \rbrace_j
        -
        \lbrace y \rbrace_j
        \right\vert
        = e^{x_{\max}} \Vert x - y \Vert_1.
      \end{displaymath}

    \item
      The linear bounds \eqref{eq:expbound2} can be derived as the tangent line in $\lbrace x \rbrace_i = 0$ (in the case of lower bound) and the secant line connecting points $[0,e^0]$ and $[1,e^1]$ (upper bound). Summing inequalities \eqref{eq:expbound2} through $i = 1,\dots,K$ and using the property of simplex $\sum\limits_{i=1}^K \lbrace x \rbrace_i = 1$ gives bounds \eqref{eq:expsumbound}.

  \end{itemize}
\end{proof}

\begin{lemma}[$l^1$ Lipschitz constant of softmax on bounded domain]
  \label{lemma:L1softmax_bounded}
  For any $x,y \in [x_{\min},x_{\max}]^K \subset \mathbb{R}^K$, we have
  \begin{equation}
    \Vert \softmax (x) - \softmax (y) \Vert_1
    \leq L^{\langle 1 \rangle}_K \Vert x - y\Vert_1
  \end{equation}
  with
  \begin{equation}
    L^{\langle 1 \rangle}_K \leq \frac{2e^{x_{\max}}}{K e^{x_{\min}}}.
  \end{equation}
\end{lemma}

\begin{proof}
  Using the definition of softmax function and $l^1$ norm, we can write
  \begin{equation}
    \label{eq:proof3}
    \Vert \softmax (x) - \softmax (y) \Vert_1
    =
    \sum\limits_{i=1}^K
    \left\vert
    \frac{e^{\lbrace x \rbrace_i}}{\sum\limits_{j=1}^K e^{\lbrace x \rbrace_j}}
    -
    \frac{e^{\lbrace y \rbrace_i}}{\sum\limits_{j=1}^K e^{\lbrace y \rbrace_j}}
    \right\vert
    .
  \end{equation}
  Each term of this sum can be estimated using Lemma \ref{lemma:absest}
  \begin{equation}
    \label{eq:proof2}
    \sum\limits_{i=1}^K
    \left\vert
    \frac{e^{\lbrace x \rbrace_i}}{\sum\limits_{j=1}^K e^{\lbrace x \rbrace_j}}
    -
    \frac{e^{\lbrace y \rbrace_i}}{\sum\limits_{j=1}^K e^{\lbrace y \rbrace_j}}
    \right\vert
    \leq
    \sum\limits_{i=1}^K
    \left(
      \underbrace{
        \frac{\vert e^{\lbrace x \rbrace_i} - e^{\lbrace y \rbrace_i} \vert}{\sum\limits_{j=1}^K e^{\lbrace x \rbrace_j}}
      }_{=: A_i}
      +
      \underbrace{
        \frac{e^{\lbrace y \rbrace_i}}{\left( \sum\limits_{j=1}^K e^{\lbrace x \rbrace_j} \right) \left( \sum\limits_{j=1}^K e^{\lbrace y \rbrace_j} \right)}
        \left\vert
        \sum\limits_{j=1}^K e^{\lbrace x \rbrace_j} - \sum\limits_{j=1}^K e^{\lbrace y \rbrace_j}
        \right\vert
      }_{=: B_i}
    \right)
  \end{equation}
  We further estimate each term:
  \begin{enumerate}
    \item[$A_i$] Using the lower bound in \eqref{eq:expsumbound_bound} and upper bound in \eqref{eq:expL}
      \begin{displaymath}
        \frac{\vert e^{\lbrace x \rbrace_i} - e^{\lbrace y \rbrace_i} \vert}{\sum\limits_{j=1}^K e^{\lbrace x \rbrace_j}}
        \leq
        \frac{1}{Ke^{x_{\min}}} \vert e^{\lbrace x \rbrace_i} - e^{\lbrace y \rbrace_i} \vert
        \leq
        \frac{e^{x_{\max}}}{Ke^{x_{\min}}} \vert \lbrace x \rbrace_i - \lbrace y \rbrace_i \vert
        .
      \end{displaymath}

    \item[$B_i$] Using the lower bound in \eqref{eq:expsumbound} and upper bound \eqref{eq:expsumL}
      \begin{displaymath}
        \frac{e^{\lbrace y \rbrace_i}}{\left( \sum\limits_{j=1}^K e^{\lbrace x \rbrace_j} \right) \left( \sum\limits_{j=1}^K e^{\lbrace y \rbrace_j} \right)}
        \left\vert
        \sum\limits_{j=1}^K e^{\lbrace x \rbrace_j} - \sum\limits_{j=1}^K e^{\lbrace y \rbrace_j}
        \right\vert
        \leq
        \frac{e^{\lbrace y \rbrace_i}}{\sum\limits_{j=1}^K e^{\lbrace y \rbrace_j}}
        \frac{e^{x_{\max}}}{Ke^{x_{\min}}} \Vert x - y \Vert_1.
      \end{displaymath}
  \end{enumerate}

  \noindent Using these estimations, the upper bound \eqref{eq:proof2} becomes
  \begin{displaymath}
    \begin{array}{rcl}
      \sum\limits_{i=1}^K
      \left\vert
      \frac{e^{\lbrace x \rbrace_i}}{\sum\limits_{j=1}^K e^{\lbrace x \rbrace_j}}
      -
      \frac{e^{\lbrace y \rbrace_i}}{\sum\limits_{j=1}^K e^{\lbrace y \rbrace_j}}
      \right\vert
      & \leq &
      \sum\limits_{i=1}^K
      \left(
        \frac{e^{x_{\max}}}{Ke^{x_{\min}}} \vert \lbrace x \rbrace_i - \lbrace y \rbrace_i \vert
        +
        \frac{e^{\lbrace y \rbrace_i}}{\sum\limits_{j=1}^K e^{\lbrace y \rbrace_j}}
        \frac{e^{x_{\max}}}{Ke^{x_{\min}}} \Vert x - y \Vert_1
      \right) \\
      & &
      =
      \frac{e^{x_{\max}}}{Ke^{x_{\min}}}
      \underbrace{
        \left(
          \sum\limits_{i=1}^K \vert \lbrace x \rbrace_i - \lbrace y \rbrace_i \vert
        \right)
      }_{= \Vert x - y \Vert_1}
      +
      \frac{e^{x_{\max}}}{Ke^{x_{\min}}} \Vert x - y \Vert_1
      \underbrace{
        \left(
          \sum\limits_{i=1}^K
          \frac{e^{\lbrace y \rbrace_i}}{\sum\limits_{j=1}^K e^{\lbrace y \rbrace_j}}
      \right)}_{=1} \\
      & & = \frac{2e^{x_{\max}}}{Ke^{x_{\min}}} \Vert x - y \Vert_1 .
    \end{array}
  \end{displaymath}

\end{proof}

\begin{lemma}[$l^1$ Lipschitz constant of softmax on simplex]
  \label{lemma:L1softmax_simplex}
  For any $x,y \in \mathbb{P}^K$, we have
  \begin{equation}
    \Vert \softmax (x) - \softmax (y) \Vert_1
    \leq L^{\langle 1 \rangle}_K \Vert x - y\Vert_1
  \end{equation}
  with
  \begin{equation}
    L^{\langle 1 \rangle}_K \leq \frac{2e}{K + 1}.
  \end{equation}
\end{lemma}

\begin{proof}
  We use the definition of softmax function and use the inequality provided by Lemma \ref{lemma:absest} to estimate each term in \eqref{eq:proof3} to obtain \eqref{eq:proof2}.

  However, comparing to the proof of Lemma \ref{lemma:L1softmax_bounded}, we estimate $A_i$ and $B_i$ differently:

  \begin{enumerate}
    \item[$A_i$] Using the lower bound in \eqref{eq:expsumbound} and upper bound in \eqref{eq:expL}
      \begin{displaymath}
        \frac{\vert e^{\lbrace x \rbrace_i} - e^{\lbrace y \rbrace_i} \vert}{\sum\limits_{j=1}^K e^{\lbrace x \rbrace_j}}
        \leq
        \frac{1}{K+1} \vert e^{\lbrace x \rbrace_i} - e^{\lbrace y \rbrace_i} \vert
        \leq
        \frac{e}{K+1} \vert \lbrace x \rbrace_i - \lbrace y \rbrace_i \vert
        .
      \end{displaymath}

    \item[$B_i$] Using the lower bound in \eqref{eq:expsumbound_bound} and upper bound \eqref{eq:expsumL}
      \begin{displaymath}
        \frac{e^{\lbrace y \rbrace_i}}{\left( \sum\limits_{j=1}^K e^{\lbrace x \rbrace_j} \right) \left( \sum\limits_{j=1}^K e^{\lbrace y \rbrace_j} \right)}
        \left\vert
        \sum\limits_{j=1}^K e^{\lbrace x \rbrace_j} - \sum\limits_{j=1}^K e^{\lbrace y \rbrace_j}
        \right\vert
        \leq
        \frac{e^{\lbrace y \rbrace_i}}{\sum\limits_{j=1}^K e^{\lbrace y \rbrace_j}}
        \frac{e}{K+1} \Vert x - y \Vert_1.
      \end{displaymath}
  \end{enumerate}

  \noindent Using these estimations, the upper bound \eqref{eq:proof2} becomes
  \begin{displaymath}
    \begin{array}{rcl}
      \sum\limits_{i=1}^K
      \left\vert
      \frac{e^{\lbrace x \rbrace_i}}{\sum\limits_{j=1}^K e^{\lbrace x \rbrace_j}}
      -
      \frac{e^{\lbrace y \rbrace_i}}{\sum\limits_{j=1}^K e^{\lbrace y \rbrace_j}}
      \right\vert
      & \leq &
      \sum\limits_{i=1}^K
      \left(
        \frac{e}{K+1} \vert \lbrace x \rbrace_i - \lbrace y \rbrace_i \vert
        +
        \frac{e^{\lbrace y \rbrace_i}}{\sum\limits_{j=1}^K e^{\lbrace y \rbrace_j}}
        \frac{e}{K+1} \Vert x - y \Vert_1
      \right) \\
      & &
      =
      \frac{e}{K+1}
      \underbrace{
        \left(
          \sum\limits_{i=1}^K \vert \lbrace x \rbrace_i - \lbrace y \rbrace_i \vert
        \right)
      }_{= \Vert x - y \Vert_1}
      +
      \frac{e}{K+1} \Vert x - y \Vert_1
      \underbrace{
        \left(
          \sum\limits_{i=1}^K
          \frac{e^{\lbrace y \rbrace_i}}{\sum\limits_{j=1}^K e^{\lbrace y \rbrace_j}}
      \right)}_{=1} \\
      & & = \frac{2e}{K + 1} \Vert x - y \Vert_1 .
    \end{array}
  \end{displaymath}
\end{proof}

\begin{lemma}[$l^2$ Lipschitz constant of softmax]
  \label{lemma:Lipschitz_softmax}
  Let $K \geq 2$. Then
  \begin{itemize}
    \item (hypercube)
      \begin{equation}
        \label{eq:Lipschitz_softmax}
        \forall x,y \in [x_{\min},x_{\max}]^K \subset \mathbb{R}^K:
        ~~
        \Vert \softmax(x) - \softmax(y) \Vert_2
        \leq L^{\langle 2 \rangle}_K \Vert x - y \Vert_2
      \end{equation}
      with
      \begin{equation}
        \label{eq:Lipschitz_softmax_L_bound}
        L^{\langle 2 \rangle}_K \leq
        \min
        \left\lbrace
        \frac{1}{2},
        \frac{2e^{x_{\max}}}{\sqrt{K} e^{x_{\min}}}
        \right\rbrace .
      \end{equation}
    \item (simplex)
      \begin{equation}
        \label{eq:Lipschitz_softmax_simplex}
        \forall x,y \in \mathbb{P}^K:
        ~~
        \Vert \softmax(x) - \softmax(y) \Vert_2
        \leq L^{\langle 2 \rangle}_K \Vert x - y \Vert_2
      \end{equation}
      with
      \begin{equation}
        \label{eq:Lipschitz_softmax_L}
        L^{\langle 2 \rangle}_K \leq
        \min
        \left\lbrace
        \frac{1}{2},
        \frac{2e\sqrt{K}}{K + 1}
        \right\rbrace .
      \end{equation}
  \end{itemize}
\end{lemma}

\begin{proof}
  Since
  \begin{displaymath}
    \forall x\in \mathbb{R}^K: \Vert x \Vert_2 \leq \Vert x \Vert_1 \leq \sqrt{K} \Vert x \Vert_2,
  \end{displaymath}
  we can write for any $f: \mathbb{R}^K \rightarrow \mathbb{R}^{\widehat{K}}$
  \begin{displaymath}
    \Vert f(x) - f(y) \Vert_2 \leq \Vert f(x) - f(y) \Vert_1
    \leq L^{\langle 1 \rangle} \Vert x - y \Vert_1
    \leq \sqrt{K} L^{\langle 1 \rangle} \Vert x - y \Vert_2
  \end{displaymath}
  and consequently,
  \begin{displaymath}
    L^{\langle 2 \rangle} \leq \sqrt{K} L^{\langle 1 \rangle}.
  \end{displaymath}
  This inequality allows us to apply the Lemma \ref{lemma:L1softmax_bounded} and Lemma \ref{lemma:L1softmax_simplex}. The rest of the proof focuses on the constant bound.

  The Lipschitz constant of a differentiable mapping can be estimated from above by the matrix norm of Jacobian matrix corresponding to the chosen vector norm
  \begin{equation}
    \label{eq:proof_Lipchitz_est}
    L \leq \max\limits_{x \in \mathbb{R}^K} \Vert J(x) \Vert_2,
  \end{equation}
  where $J: \mathbb{R}^K \rightarrow \mathbb{R}^{K \times K}$ is the Jacobian matrix of the softmax function given by
  \begin{equation}
    \label{eq:softmax_Jacobian}
    J(x) = \diag (\softmax(x)) - \softmax(x) ~ \softmax(x)^{\mathsf{T}}.
  \end{equation}
  In the case of the Euclidean norm, the corresponding matrix norm is the spectral norm, i.e., the largest eigenvalue of symmetric positive semidefinite matrix \eqref{eq:softmax_Jacobian}.
  For given arbitrary $x \in \mathbb{R}^K$, we denote
  \begin{displaymath}
    p := \softmax(x) \in \mathbb{P}^K
  \end{displaymath}

  We solve the problem equivalent to \eqref{eq:proof_Lipchitz_est} in form
  \begin{equation}
    \label{eq:proof_Lipchitz_est2}
    L \leq
    \max\limits_{p \in \mathbb{P}^K} \Vert \widehat{J}(p) \Vert_2 ,
  \end{equation}
  where components of the matrix $\widehat{J}(p) \in \mathbb{R}^{K,K}$ are given by
  \begin{displaymath}
    \lbrace \widehat{J}(p) \rbrace_{i,j} = \left\lbrace
    \begin{array}{ll}
      \lbrace p \rbrace_i(1- \lbrace p \rbrace_i) & \textrm{if} ~ i = j,\\
      - \lbrace p \rbrace_i \lbrace p \rbrace_j & \textrm{if} ~ i \neq j.
    \end{array}
    \right.
  \end{displaymath}

  Applying the Gershgorin circle theorem, the largest eigenvalue can be estimated as the sum of the center of the circle, i.e., the diagonal element, and the radius of the circle, given by the sum of the absolute values of the off-diagonal elements. For any $i$, we obtain the inequality
  \begin{displaymath}
    \Vert \widehat{J}(p) \Vert_2 \leq \lbrace p \rbrace_i(1- \lbrace p \rbrace_i) +
    \lbrace p \rbrace_i \sum\limits_{j=1, j \neq i}^K \lbrace p \rbrace_j =
    \lbrace p \rbrace_i ( 1 - \lbrace p \rbrace_i ) +
    \lbrace p \rbrace_i (1 - \lbrace p \rbrace_i) = 2 \lbrace p \rbrace_i(1-\lbrace p \rbrace_i)
    ,
  \end{displaymath}
  which is a concave quadratic function that attains its maximum value at $\frac{1}{2}$.

\end{proof}

\begin{lemma}[$\Gamma$-problem: sufficient condition for unique solution]
  \label{lemma:Gamma_unique}
  If $S, \theta, \gamma_{(0)}$ are fixed and
  \begin{displaymath}
    \min \left\lbrace \epsilon_n ~\vert~ n = 1,\dots,N+1 \right\rbrace
    ~~ > ~~
    \max \left\lbrace \Vert A_{(n)} \Vert_2 + \Vert A_{(n+1)} \Vert_2 ~\vert~ n = 1,\dots,N \right\rbrace,
  \end{displaymath}
  where
  \begin{equation}
    \label{eq:An}
    A_{(n)} \in \mathbb{R}^{K_{n+1} \times K_n}, n = 1,\dots,N,
    ~~~
    \lbrace A_{(n)} \rbrace_{k_{n+1},k_n} :=
    - \delta_i \log( \lbrace \theta^{(n)} \rbrace_{k_{n+1},k_n}),
  \end{equation}
  then the objective function of \eqref{eq:EON_fun_E2} is strictly convex and, consequently, the solution $\hat{\Gamma}$ is unique.
\end{lemma}

\begin{proof}
  We denote the total number of remaining unknowns in $\Gamma$ by
  \begin{equation}
    \Ksum := \sum\limits_{n=1}^{N+1} K_n.
  \end{equation}
  The problem is to solve \eqref{eq:EON_fun_E2} with fixed variables $S$ and $\theta$. At first, notice that the problem is separable in $t$  and for all $t = 1,\dots,T$, we obtain vector of unknowns
  \begin{equation}
    \label{eq:gamma_block}
    \gamma_t :=
    \left[
      \begin{array}{c}
        \{ \gamma_{(1)} \}_{(:,t)}, \\
        \vdots, \\
        \{ \gamma_{(N+1)} \}_{(:,t)}
      \end{array}
    \right]
    \in
    \mathbb{R}^{\Ksum}.
  \end{equation}
  We simplify the objective function to the form
  \begin{equation}
    \label{eq:L_Gamma_problem}
    L_{t,S,\theta}(\gamma) =
    \langle b_t, \gamma_{(1)} \rangle
    +
    \sum\limits_{n=1}^{N}
    \langle \gamma_{(n+1)}, A_{(n)} \gamma_{(n)} \rangle
    +
    \sum\limits_{n=0}^{N+1}
    \epsilon_n
    \langle \gamma_{(n)}, \log \gamma_{(n)} \rangle,
  \end{equation}
  where
  \begin{equation}
    \label{eq:Gamma_bt}
    b_{t} \in \mathbb{R}^{K_1},
    ~~~
    \lbrace b_{t} \rbrace_{k_1} :=
    \sum\limits_{d=1}^{K_0} \lbrace \gamma_{(0)} \rbrace_{d,t} ~ \big( \lbrace X(t) \rbrace_{d} - S_{d,k_1} \big)^2,
  \end{equation}
  and \eqref{eq:An}.
  The feasible set corresponding to block-vector \eqref{eq:gamma_block} can be derived from \eqref{eq:Omega_E2} and is given by
  \begin{displaymath}
    \Omega_{\gamma} =
    \lbrace \gamma \in \mathbb{R}^{\Ksum}
    ~ \vert ~
    \forall n = 1,\dots,N+1:
    \gamma_{(n)} \in \mathbb{P}^{K_n}
    \rbrace.
  \end{displaymath}

  The gradient of the objective function of \eqref{eq:L_Gamma_problem} can be easily derived as
  \begin{equation}
    \label{eq:L_Gamma_gradient}
    \nabla L_{t,S,\theta}(\gamma) =
    \left[
      \begin{array}{l}
        b_t + A_{(1)}^{\mathsf{T}} \gamma_{(2)} \\[2mm]
        A_{(1)} \gamma_{(1)} + A_{(2)}^{\mathsf{T}} \gamma_{(3)} \\[2mm]
        A_{(2)} \gamma_{(2)} + A_{(3)}^{\mathsf{T}} \gamma_{(4)} \\[2mm]
        \vdots \\[2mm]
        A_{(N-1)} \gamma_{(N-1)} + A_{(N)}^{\mathsf{T}} \gamma_{(N+1)} \\[2mm]
        A_{(N)} \gamma_{(N)}
      \end{array}
    \right]
    +
    \left[
      \begin{array}{l}
        \epsilon_1 (\log \gamma_{(1)} + \onebb_{K_1}) \\[2mm]
        \epsilon_2 (\log \gamma_{(2)} + \onebb_{K_2}) \\[2mm]
        \epsilon_3 (\log \gamma_{(3)} + \onebb_{K_3}) \\[2mm]
        \vdots \\[2mm]
        \epsilon_{N}  (\log \gamma_{(N)} + \onebb_{K_N}) \\[2mm]
        \epsilon_{N+1} (\log \gamma_{(N+1)} + \onebb_{K_{N+1}})
      \end{array}
    \right],
  \end{equation}
  and the Hessian matrix
  \begin{equation}
    \label{eq:L_Gamma_Hessian}
    \nabla^2 L_{t,S,\theta}(\gamma) =
    \underbrace{
      \left[
        \begin{array}{cccccc}
          0 & A_{(1)}^{\mathsf{T}} & & & & \\[2mm]
          A_{(1)} & 0 & A_{(2)}^{\mathsf{T}} & & & \\[2mm]
          & A_{(2)} & 0 & A_{(3)}^{\mathsf{T}} & & \\[2mm]
          & & \ddots & \ddots & \ddots \\[2mm]
          & & & A_{(N-1)} & 0 & A_{(N)}^{\mathsf{T}} \\[2mm]
          & & & & A_{(N)} & 0
        \end{array}
    \right]}_{=: A}
    +
    \diag_{n = 1,\dots,N+1} \left( \epsilon_n \diag_{k_n = 1,\dots,K_n} \left( \frac{1}{ \lbrace \gamma_{(n)} \rbrace_{k_n} } \right) \right).
  \end{equation}
  At first, notice that this matrix is real and symmetric, therefore it has real spectrum.
  The matrix $A \in \mathbb{R}^{\Ksum \times \Ksum}$ is the Hessian of the bilinear form and the second term can be interpreted as a regularization part. Similarly to well-known Tikhonov regularization, we show that if this regularization is sufficient, then the spectrum of the first term is sufficiently shifted such that the overall matrix has the smallest eigenvalue larger than zero, therefore it is positive definite and the objective function is strictly convex.

  To estimate the smallest eigenvalue of the first term, we apply Generalized Gershgorin Circle theorem for block matrices. Based on the zero diagonal blocks, the centers of circles are located at zeros. The radius of each circle is given by the sum of the 2-norm of off-diagonal blocks in each row, i.e., the spectral radii of these blocks. The first and the last row includes only one non-zero block, however, these blocks are included also in other rows, therefore the radius of these blocks will be always smaller then radius of rows which includes these blocks with additional blocks. The remaining rows include two non-zero blocks, therefore the largest radius of all blocks can be estimated as
  \begin{displaymath}
    R_{\max} = \max \left\lbrace \Vert A_{(n)} \Vert_2 + \Vert A_{(n+1)} \Vert_2 ~\vert~ n = 1,\dots,N \right\rbrace.
  \end{displaymath}
  Since centers are located in zero and all eigenvalues are real, the spectrum of the first part can be located in the interval
  \begin{displaymath}
    \lambda ~ \in ~ \left[ -R_{\max}, R_{\max} \right].
  \end{displaymath}
  The idea is to shift this interval to strictly positive numbers, i.e., add to spectrum value larger than $R_{\max}$. This will be performed by the shift of centers of Gershgorin circles, i.e., by the shift of diagonal of overall Hessian matrix. And this is achieved using the second term in \eqref{eq:L_Gamma_Hessian}.

  We are interested in the properties of objective function (i.e., the Hessian matrix) only on the feasible set of the problem \eqref{eq:L_Gamma_problem}. In this case, $\gamma_{(n)} \in \mathbb{P}^{K_n}$ and, consequently, diagonal elements of the regularization part can be estimated by
  \begin{displaymath}
    \forall n,k_n: ~~\frac{1}{\lbrace \gamma_{(n)} \rbrace_{k_n}} \geq 1.
  \end{displaymath}
  In the worst case scenario, it is necessary to scale the lowest value $1$ by coefficient $\epsilon_n$ such that the smallest value in diagonal will be larger than $R_{\max}$. Therefore, all regularization parameters $\epsilon_n$ have to be strictly larger than $R_{\max}$. This gives us the sufficient condition on regularization parameters presented in the lemma.
\end{proof}

\begin{lemma}[$\Gamma$-problem: equivalence with the system of softmax equations]
  \label{lemma:Gamma_blocksoftmax_solution}
  The solution of $\Gamma$ problem (i.e., problem \eqref{eq:EON_fun_E2} with fixed variables $S, \theta$, and $\gamma_{(0)}$) is equivalent to the solution of the system
  \begin{equation}
    \label{eq:gamma_softmaxeq}
    \gamma = \underset{K_1,\dots,K_{N+1}}{\softmax} (-E^{-1} (A \gamma + b))
  \end{equation}
  where the structure of variable $\gamma$ is given by \eqref{eq:gamma_block}, the matrix $A \in \mathbb{R}^{\Ksum \times \Ksum}$ is given by \eqref{eq:L_Gamma_Hessian}, $b^{\mathsf{T}} := [b_t^{\mathsf{T}},0, \dots,0]  \in \mathbb{R}^{\Ksum}$ with $b_t$ defined by \eqref{eq:Gamma_bt}, and matrix $E \in \mathbb{R}^{\Ksum \times \Ksum}$ is a block diagonal matrix of regularization coefficients
  \begin{equation}
    \label{eq:Gamma_proof_E}
    E := \diag_{n = 1,\dots,N+1} \left( \epsilon_n I_{K_n} \right).
  \end{equation}
\end{lemma}

\begin{proof}
  In the derivation, we ignore inequality constraints, however, the derived solution satisfies the inequality constraints. We introduce the Lagrange function
  \begin{displaymath}
    \mathcal{L} (\gamma,\mu) := L_{t,S,\theta}(\gamma) +
    \sum\limits_{n=1}^{N+1}
    \mu_n (\onebb^{\mathsf{T}}_{K_n} \gamma_{(n)}  - 1),
  \end{displaymath}
  where $L_{t,S,\theta}$ is given by \eqref{eq:L_Gamma_problem} and $\mu \in \mathbb{R}^{N+1}$ is a vector of Lagrange multipliers corresponding to equality constraints.

  The Karush-Kuhn-Tucker optimality conditions are given by the derivatives with respect to primal variables
  \begin{equation}
    \label{eq:Gamma_KKT1}
    \begin{array}{rclllcl}
      \nabla_{\gamma_{(1)}} \mathcal{L} (\gamma,\mu)
      & = & b_t + A_{(1)}^{\mathsf{T}} \gamma_{(2)} & +~ \epsilon_1 (\log \gamma_{(1)} + \onebb_{K_1}) & +~ \mu_1 \onebb_{K_1}
      & = & \zerobb_{K_1} \\[2mm]
      \nabla_{\gamma_{(2)}} \mathcal{L} (\gamma,\mu)
      & = & A_{(1)} \gamma_{(1)} + A_{(2)}^{\mathsf{T}} \gamma_{(3)} & +~ \epsilon_2 (\log \gamma_{(2)} + \onebb_{K_2}) & +~ \mu_2 \onebb_{K_2}
      & = & \zerobb_{K_2} \\[2mm]
      \nabla_{\gamma_{(3)}} \mathcal{L} (\gamma,\mu)
      & = & A_{(2)} \gamma_{(2)} + A_{(3)}^{\mathsf{T}} \gamma_{(4)} & +~ \epsilon_3 (\log \gamma_{(3)} + \onebb_{K_3}) & +~ \mu_3 \onebb_{K_3}
      & = & \zerobb_{K_3} \\[2mm]
      & \vdots & & & \\[2mm]
      \nabla_{\gamma_{(N)}} \mathcal{L} (\gamma,\mu)
      & = & A_{(N-1)} \gamma_{(N-1)} + A_{(N)}^{\mathsf{T}} \gamma_{(N+1)} & +~ \epsilon_N (\log \gamma_{(N)} + \onebb_{K_N}) & +~ \mu_N \onebb_{K_N}
      & = & \zerobb_{K_N} \\[2mm]
      \nabla_{\gamma_{(N+1)}} \mathcal{L} (\gamma,\mu)
      & = & A_{(N)} \gamma_{(N)} &  +~ \epsilon_{N+1} (\log \gamma_{(N+1)} + \onebb_{K_{N+1}}) & +~ \mu_{N+1} \onebb_{K_{N+1}}
      & = & \zerobb_{K_{N+1}}
    \end{array}
  \end{equation}
  and derivatives with respect to dual variables
  \begin{equation}
    \label{eq:Gamma_KKT2}
    \forall n = 1,\dots,N+1:
    ~~~
    \frac{\partial \mathcal{L} }{\partial \mu_n} (\gamma,\mu)
    = \onebb^{\mathsf{T}}_{K_n} \gamma_{(n)}  - 1 = 0 .
  \end{equation}

  The system \eqref{eq:Gamma_KKT1} can be rewritten as
  \begin{equation}
    \label{eq:Gamma_KKT1b}
    \begin{array}{rcll}
      \log \gamma_{(1)}
      & = & -\epsilon_1^{-1}(b_t + A_{(1)}^{\mathsf{T}} \gamma_{(2)})
      & -~ (\onebb_{K_1} + \epsilon_1^{-1} \mu_1 \onebb_{K_1})\\ [2mm]
      \log \gamma_{(2)}
      & = & -\epsilon_2^{-1}(A_{(1)} \gamma_{(1)} + A_{(2)}^{\mathsf{T}} \gamma_{(3)})
      & -~ (\onebb_{K_2} + \epsilon_2^{-1}\mu_2 \onebb_{K_2} ) \\ [2mm]
      \log \gamma_{(3)}
      & = & -\epsilon_3^{-1}(A_{(2)} \gamma_{(2)} + A_{(3)}^{\mathsf{T}} \gamma_{(4)})
      & -~ (\onebb_{K_3} + \epsilon_3^{-1}\mu_3 \onebb_{K_3} ) \\ [2mm]
      & \vdots & \\[2mm]
      \log \gamma_{(N)}
      & = & -\epsilon_N^{-1}(A_{(N-1)} \gamma_{(N-1)} + A_{(N)}^{\mathsf{T}} \gamma_{(N+1)})
      & -~ (\onebb_{K_3} + \epsilon_N^{-1}\mu_N \onebb_{K_N} ) \\ [2mm]
      \log \gamma_{(N+1)}
      & = & -\epsilon_{N+1}^{-1}(A_{(N)} \gamma_{(N)})
      & -~ (\onebb_{K_{N+1}} + \epsilon_{N+1}^{-1}\mu_{N+1} \onebb_{K_{N+1}} )
    \end{array}
  \end{equation}
  For the simplicity, we denote the first term in each equation of \eqref{eq:Gamma_KKT1b} by vector $q_{(n)} \in \mathbb{R}^{K_n}$ defined as
  \begin{equation}
    \label{eq:Gamma_KKT1bdef}
    q_{(n)} =
    \left\lbrace
    \begin{array}{ll}
      -\epsilon_1^{-1}(b_t + A_{(1)}^{\mathsf{T}} \gamma_{(2)}) &~ \mathrm{for}~ n = 1, \\[2mm]
      -\epsilon_n^{-1}(A_{(n-1)} \gamma_{(n-1)} + A_{(n)}^{\mathsf{T}} \gamma_{(n+1)}) &~ \mathrm{for}~ 1 < n < N+1, \\[2mm]
      -\epsilon_{N+1}^{-1}(A_{(N)} \gamma_{(N)}) &~ \mathrm{for}~ n = N+1.
    \end{array}
    \right.
  \end{equation}
  and we can write system \eqref{eq:Gamma_KKT1b} in the compact form
  \begin{equation}
    \label{eq:Gamma_KKT1c}
    \log \gamma_{(n)}
    = q_{(n)} - (\onebb_{K_n} + \epsilon_n^{-1}\mu_n \onebb_{K_n} ),
    ~~~ n = 1,\dots,N+1.
  \end{equation}
  After the application of exponential function on both sides, we obtain for $k_n$-th component of vector $\gamma_{(n)}$ the equality
  \begin{equation}
    \label{eq:Gamma_KKT1d}
    \forall k_n = 1,\dots,K_n: ~~~ \lbrace\gamma_{(n)}\rbrace_{k_n} = \frac{\exp \lbrace q_{(n)}\rbrace_{k_n}}{\exp (1 + \epsilon_n^{-1}\mu_n)},
  \end{equation}
  which can be substituted into equality constraint \eqref{eq:Gamma_KKT2} to get
  \begin{displaymath}
    \sum\limits_{k_n=1}^{K_n} \frac{\exp \lbrace q_{(n)}\rbrace_{k_n}}{\exp (1 + \epsilon_n^{-1}\mu_n)} = 1
    ~~~~
    \Rightarrow
    ~~~~
    \exp (1 + \epsilon_n^{-1}\mu_n) = \sum\limits_{k_n=1}^{K_n} \exp \lbrace q_{(n)} \rbrace_{k_n}
  \end{displaymath}
  which can be substituted back into \eqref{eq:Gamma_KKT1d} to obtain
  \begin{equation}
    \label{eq:Gamma_KKT1e}
    \forall k_n = 1,\dots,K_n: ~~~ \lbrace \gamma_{(n)} \rbrace_{k_n} = \frac{\exp \lbrace q_{(n)} \rbrace_{k_n}}{\sum\limits_{\hat{k}_n=1}^{K_n} \exp \lbrace q_{(n)} \rbrace_{\hat{k}_n} }.
  \end{equation}
  The function on right-hand side is the well-known $\softmax$ function, therefore, we can rewrite the system of equations \eqref{eq:Gamma_KKT1e} with \eqref{eq:Gamma_KKT1bdef} as
  \begin{equation}
    \label{eq:Gamma_KKT1f}
    \begin{array}{rcl}
      \gamma_{(1)}  & = & \softmax (-\epsilon_1^{-1}(b_t + A_{(1)}^{\mathsf{T}} \gamma_{(2)})) \\[2mm]
      \gamma_{(2)}  & = & \softmax (-\epsilon_2^{-1}(A_{(1)} \gamma_{(1)} + A_{(2)}^{\mathsf{T}} \gamma_{(3)})) \\[2mm]
      \gamma_{(3)}  & = & \softmax (-\epsilon_3^{-1}(A_{(2)} \gamma_{(2)} + A_{(3)}^{\mathsf{T}} \gamma_{(4)})) \\[2mm]
      & \vdots & \\[2mm]
      \gamma_{(N)}  & = & \softmax (-\epsilon_N^{-1}(A_{(N-1)} \gamma_{(N-1)} + A_{(N)}^{\mathsf{T}} \gamma_{(N+1)})) \\[2mm]
      \gamma_{(N+1)}  & = & \softmax (-\epsilon_{N+1}^{-1}(A_{(N)} \gamma_{(N)})),
    \end{array}
  \end{equation}
  which is the equation presented in the lemma. To conclude the proof, note that the solution of this system satisfies the inequality constraints imposed by the softmax function, which ensures non-negative values.
\end{proof}

\begin{lemma}[$\Gamma$-problem: coordinate descent algorithm convergence]
  \label{lemma:Gamma_subspace_algorithm}
  Let $S, \theta$, and $\gamma_{(0)}$ in the problem \eqref{eq:EON_fun_E2} be fixed and let the structure of variable $\gamma$ be given by \eqref{eq:gamma_block}.
  Let $\gamma^0 \in \Omega_{\gamma}$ denote the initial approximation. The sequence generated by coordinate descent algorithm \eqref{eq:generalsubspace} which solves $\gamma$ in layer-by-layer approach from the last layer to the first one is given by (for iterations $\iit = 0,1,2,\dots$)
  \begin{equation}
    \label{eq:layerbylayer}
    \begin{array}{rcl}
      \gamma_{(N+1)}^{\iit+1} & = & \arg \min\limits_{\gamma_{(N+1)} \in \mathbb{P}^{K_{N+1}}} L_{t,S,\theta}([\gamma_{(1)}^{\iit},\gamma_{(2)}^{\iit},\dots,\gamma_{(N)}^{\iit},\gamma_{(N+1)}]), \\[2mm]
      \gamma_{(N)}^{\iit+1} & = & \arg \min\limits_{\gamma_{(N)} \in \mathbb{P}^{K_N}} L_{t,S,\theta}([\gamma_{(1)}^{\iit},\gamma_{(2)}^{\iit},\dots,\gamma_{(N)},\gamma_{(N+1)}^{\iit + 1}]), \\[2mm]
      & \vdots & \\
      \gamma_{(2)}^{\iit+1} & = & \arg \min\limits_{\gamma_{(2)} \in \mathbb{P}^{K_2}} L_{t,S,\theta}([\gamma_{(1)}^{\iit},\gamma_{(2)},\dots,\gamma_{(N)}^{\iit + 1},\gamma_{(N+1)}^{\iit + 1}]), \\[2mm]
      \gamma_{(1)}^{\iit+1} & = & \arg \min\limits_{\gamma_{(1)} \in \mathbb{P}^{K_1}} L_{t,S,\theta}([\gamma_{(1)},\gamma_{(2)}^{\iit + 1},\dots,\gamma_{(N)}^{\iit + 1},\gamma_{(N+1)}^{\iit + 1}]).
    \end{array}
  \end{equation}
  The computational complexity of one iteration of this algorithm is $\mathcal{O}(T K_{\textrm{all}})$.

  Additionally, if
  \begin{equation}
    \label{eq:layerbylayer_condition}
    L_G
    +
    L_H
    <
    \frac{1}{L_{\mathrm{bsf}}},
    ~
    L_{\mathrm{bsf}} := \max\limits_{n} \left\lbrace L^{\langle 2 \rangle}_{K_n} \right\rbrace,
    ~ L_G := \max\limits_{n} \lbrace \epsilon_n^{-1} \Vert A_{(n)} \Vert_2 \rbrace,
    ~ L_H := \max\limits_{n} \lbrace \epsilon_{n+1}^{-1} \Vert A_{(n)} \Vert_2 \rbrace,
  \end{equation}
  the convergence of sequence \eqref{eq:layerbylayer} to the solution $\hat{\gamma}$ of the problem \eqref{eq:gamma_softmaxeq} can be estimated as
  \begin{equation}
    \label{eq:layerbylayer_convergence}
    \Vert \gamma^{\iit + 1} - \hat{\gamma} \Vert_2 \leq
    \frac{(\tilde{L})^{\iit}}{1 - \tilde{L}} \Vert \gamma^{\mathrm{0}} - \gamma^{\mathrm{1}} \Vert_2
  \end{equation}
  with the estimation of Lipchitz constant
  \begin{equation}
    \label{eq:layerbylayer_Lest}
    \tilde{L} := \frac{L_{\mathrm{bsf}} L_H}{1 - L_{\mathrm{bsf}} L_G}  \in (0,1).
  \end{equation}
\end{lemma}

\begin{proof}
  Because of Lemma \ref{lemma:LP_with_entropy}, the iteration process \eqref{eq:layerbylayer} can be written as a sequence of softmax functions
  \begin{equation}
    \label{eq:layerbylayer_softmax}
    \begin{array}{rcl}
      \gamma_{(N+1)}^{\iit+1} & = & \softmax (-\epsilon_{N+1} A_{(N)} \gamma_{(N)}^{\iit}), \\[2mm]
      \gamma_{(N)}^{\iit+1} & = & \softmax (-\epsilon_{N} ( A_{(N-1)} \gamma_{(N-1)}^{\iit} + A_{(N)}^{\mathsf{T}} \gamma_{N+1}^{\iit + 1 } )), \\[2mm]
      & \vdots & \\[2mm]
      \gamma_{(2)}^{\iit+1} & = & \softmax (-\epsilon_{2} (A_{(1)} \gamma_{(1)}^{\iit} + A_{(2)}^{\mathsf{T}} \gamma_{(3)}^{\iit + 1}) ), \\[2mm]
      \gamma_{(1)}^{\iit+1} & = & \softmax (-\epsilon_{1} (b_t + A_{(1)}^{\mathsf{T}} \gamma_{(2)}^{\iit + 1}) ).
    \end{array}
  \end{equation}
  Sequential evaluation of these equations yields the resulting computational complexity.

  Using the notation presented in Lemma \ref{lemma:Gamma_blocksoftmax_solution}, this process can be written in form
  \begin{equation}
    \label{eq:Gamma_LBL}
    \gamma^{\iit+1} = \underset{K_1,\dots,K_{N+1}}{\softmax} (-E^{-1} (G \gamma^{\iit + 1} + H \gamma^{\iit} + b))
  \end{equation}
  with
  \begin{equation}
    \label{eq:Gamma_proof_GH}
    G :=
    \left[
      \begin{array}{cccccc}
        0 & A_{(1)}^{\mathsf{T}} & & & & \\[2mm]
        & 0 & A_{(2)}^{\mathsf{T}} & & & \\[2mm]
        &  & 0 & A_{(3)}^{\mathsf{T}} & & \\[2mm]
        & & & \ddots & \ddots \\[2mm]
        & & & & 0 & A_{(N)}^{\mathsf{T}} \\[2mm]
        & & & & & 0
      \end{array}
    \right], ~~~
    H :=
    \left[
      \begin{array}{cccccc}
        0 &  & & & & \\[2mm]
        A_{(1)} & 0 &  & & & \\[2mm]
        & A_{(2)} & 0 &  & & \\[2mm]
        & & \ddots & \ddots & \\[2mm]
        & & & A_{(N-1)} & 0 &  \\[2mm]
        & & & & A_{(N)} & 0
      \end{array}
    \right].
  \end{equation}
  The equation \eqref{eq:Gamma_LBL} can be considered to be implicit fixed-point iteration scheme. It is easy to see that the solution of \eqref{eq:gamma_softmaxeq} is a fixed-point of this mapping since
  \begin{displaymath}
    A = G + H,
  \end{displaymath}
  therefore \eqref{eq:Gamma_LBL} is consistent with \eqref{eq:gamma_softmaxeq}.

  To prove and estimate convergence, we demonstrate the contractivity property. In our estimation, we use the contractivity of block softmax function. Using the contractivity given by Lemma \ref{lemma:Lipschitz_softmax},
  we can estimate (for any $v,q \in \mathbb{R}^{\Ksum}$)
  \begin{displaymath}
    \begin{array}{rcl}
      \Vert \underset{K_1,\dots,K_N}{\softmax} (v) - \underset{K_1,\dots,K_N}{\softmax} (q) \Vert_2^2
      & = & \sum\limits_{k=1}^{\Ksum}
      \left(
        \lbrace \underset{K_1,\dots,K_N}{\softmax} (v) \rbrace_k
        -
        \lbrace \underset{K_1,\dots,K_N}{\softmax} (q) \rbrace_k
      \right)^2 \\[2mm]
      & = &
      \sum\limits_{n=1}^N \Vert \softmax (v_{(n)}) - \softmax (q_{(n)}) \Vert_2^2 \leq \sum\limits_{n=1}^N \left( L_{K_n}^{\langle 2 \rangle} \right)^2 \Vert v_{(n)} - q_{(n)} \Vert_2^2 \\[2mm]
      & \leq & \max\limits_{n} \left\lbrace \left( L_{K_n}^{\langle 2 \rangle} \right)^2 \right\rbrace ~ \sum\limits_{n=1}^N \Vert v_{(n)} - q_{(n)} \Vert_2^2 = \left( \max\limits_{n} \left\lbrace  L_{K_n}^{\langle 2 \rangle}  \right\rbrace \right)^2 \Vert v - q \Vert_2^2
    \end{array}
  \end{displaymath}
  therefore
  \begin{equation}
    \label{eq:bsf_contractivity}
    \forall v,q \in \mathbb{R}^{\Ksum}: ~
    \Vert \underset{K_1,\dots,K_N}{\softmax} (v) - \underset{K_1,\dots,K_N}{\softmax} (q) \Vert_2
    \leq
    L_{\mathrm{bsf}}
    \Vert v - q \Vert_2, ~~ \textrm{with}~ L_{\mathrm{bsf}} :=  \max\limits_{n} \left\lbrace L_{K_n}^{\langle 2 \rangle} \right\rbrace.
  \end{equation}

  Let $\gamma$ and $\sigma$ be two sequences generated by \eqref{eq:Gamma_LBL}. We show that there exists contractivity constant $L < 1$ such that
  \begin{equation}
    \label{eq:gamma_contractivity}
    \Vert \gamma^{\iit + 1} - \sigma^{\iit + 1} \Vert_2 \leq L \Vert \gamma^{\iit} - \sigma^{\iit} \Vert_2.
  \end{equation}

  We start with the left hand side of \eqref{eq:gamma_contractivity}, apply the definition of mapping \eqref{eq:Gamma_LBL}, and estimate using \eqref{eq:bsf_contractivity}
  \begin{displaymath}
    \begin{array}{rcl}
      \Vert \gamma^{\iit + 1} - \sigma^{\iit + 1} \Vert_2 & = &
      \Vert
      \underset{K_1,\dots,K_{N+1}}{\softmax} (-E^{-1} (G \gamma^{\iit + 1} + H \gamma^{\iit} + b))
      -
      \underset{K_1,\dots,K_{N+1}}{\softmax} (-E^{-1} (G \sigma^{\iit + 1} + H \sigma^{\iit} + b))
      \Vert_2\\[4mm]
      & \leq & L_{\mathrm{bsf}} \Vert -E^{-1} (G \gamma^{\iit + 1} + H \gamma^{\iit} - G \sigma^{\iit + 1} - H \sigma^{\iit}) \Vert_2 \\[4mm]
      & \leq & L_{\mathrm{bsf}}
      \Vert
      E^{-1} G (\gamma^{\iit + 1} - \sigma^{\iit + 1})
      + E^{-1} H (\gamma^{\iit} - \sigma^{\iit})
      \Vert_2
    \end{array}
  \end{displaymath}

  Using the triangle inequality (i.e., $\forall x,y \in \mathbb{R}^n: \Vert x + y \Vert \leq \Vert x \Vert + \Vert y \Vert$) and inequality of matrix norm corresponding to vector norm (i.e., $\forall x \in \mathbb{R}^n \forall A \in \mathbb{R}^{n \times n}: \Vert Ax \Vert_2 \leq \Vert A \Vert_2 \Vert x \Vert_2$) we can continue the estimation

  \begin{displaymath}
    \begin{array}{rcl}
      \Vert \gamma^{\iit + 1} - \sigma^{\iit + 1} \Vert_2 & \leq &
      L_{\mathrm{bsf}} \left(
        \Vert E^{-1} G (\gamma^{\iit + 1} - \sigma^{\iit + 1}) \Vert_2
      + \Vert E^{-1} H (\gamma^{\iit} - \sigma^{\iit}) \Vert_2 \right) \\[2mm]
      & \leq & L_{\mathrm{bsf}}
      \Vert E^{-1} G \Vert_2 \Vert \gamma^{\iit + 1} - \sigma^{\iit + 1} \Vert_2
      + L_{\mathrm{bsf}} \Vert E^{-1} H \Vert_2 \Vert \gamma^{\iit} - \sigma^{\iit} \Vert_2,
    \end{array}
  \end{displaymath}

  which is equivalent to

  \begin{displaymath}
    \left( 1 -L_{\mathrm{bsf}}
    \Vert E^{-1} G \Vert_2 \right) \Vert \gamma^{\iit + 1} - \sigma^{\iit + 1} \Vert_2
    \leq
    L_{\mathrm{bsf}} \Vert E^{-1} H \Vert_2 \Vert \gamma^{\iit} - \sigma^{\iit} \Vert_2
  \end{displaymath}

  If we assume that
  \begin{equation}
    \label{eq:Gamma_proof_aux_ineq}
    1 -L_{\mathrm{bsf}} \Vert E^{-1} G \Vert_2 > 0,
  \end{equation}
  then we get inequality
  \begin{equation}
    \label{eq:Gamma_proof_aux_ineq4}
    \Vert \gamma^{\iit + 1} - \sigma^{\iit + 1} \Vert_2
    \leq
    L \Vert \gamma^{\iit} - \sigma^{\iit} \Vert_2, ~~~ \textrm{with} ~
    L:= \frac{L_{\mathrm{bsf}} \Vert E^{-1} H \Vert_2}{
      1 -L_{\mathrm{bsf}}
      \Vert E^{-1} G \Vert_2
    }.
  \end{equation}

  To derive the condition under which the Lipchitz constant $L$ is less than $1$, we solve inequality $L < 1$, which after simple manipulations (with assumption \eqref{eq:Gamma_proof_aux_ineq}) leads to condition
  \begin{equation}
    \label{eq:Gamma_proof_aux_ineq2}
    \Vert E^{-1} G \Vert_2 + \Vert E^{-1} H \Vert_2 < \frac{1}{L_{\mathrm{bsf}}}
    .
  \end{equation}
  If the condition \eqref{eq:Gamma_proof_aux_ineq2} is satisfied, then also condition \eqref{eq:Gamma_proof_aux_ineq} is satisfied since $\Vert E^{-1} H \Vert_2 \geq 0$.
  Norms in \eqref{eq:Gamma_proof_aux_ineq2} can be easily estimated using the structure of matrices $G,H$ \eqref{eq:Gamma_proof_GH}, structure of matrix $E$ \eqref{eq:Gamma_proof_E}, and block Gersghorin circle theorem
  \begin{equation}
    \label{eq:Gamma_proof_aux_ineq3}
    \Vert E^{-1} G \Vert_2 \leq \max\limits_{n} \lbrace \epsilon_n^{-1} \Vert A_{(n)} \Vert_2 \rbrace =: L_G,
    ~~~~
    \Vert E^{-1} H \Vert_2 \leq \max\limits_{n} \lbrace \epsilon_{n+1}^{-1} \Vert A_{(n)} \Vert_2 \rbrace =: L_H.
  \end{equation}
  These estimations with \eqref{eq:Gamma_proof_aux_ineq2} give the sufficient condition \eqref{eq:layerbylayer_condition} presented in the lemma.

  Because of inequalities \eqref{eq:Gamma_proof_aux_ineq3}, we have the inequality between \eqref{eq:Gamma_proof_aux_ineq4} and \eqref{eq:layerbylayer_Lest} in form
  \begin{displaymath}
    L :=
    \frac{L_{\mathrm{bsf}} \Vert E^{-1} H \Vert_2}{
      1 -L_{\mathrm{bsf}}
      \Vert E^{-1} G \Vert_2
    } \leq
    \frac{L_{\mathrm{bsf}} L_H}{1 - L_{\mathrm{bsf}} L_G}
    =: \tilde{L}.
  \end{displaymath}
  Additionally, because of \eqref{eq:layerbylayer_condition}, we have $L < 1$.

  The estimation of the speed of convergence \eqref{eq:layerbylayer_convergence} is a standard result of Banach fixed-point theorem, see, e.g., \cite{latifBanachContractionPrinciple2014}.
\end{proof}

\begin{lemma}[Coordinate descent solution for rank-1 factorization of $\gamma_{(0)}$]
  \label{lemma:gamma0_case3}
  Let $S, \theta, \gamma_{(1)}, \dots, \gamma_{(N+1)}$ in the problem \eqref{eq:EON_fun_E2} be fixed. We consider a rank-1 factorization of unknown probability matrix $\gamma_{0} \in \mathbb{R}^{K_0 \times T}$ in form
  \begin{equation}
    \label{eq:gamma0_case3}
    \gamma_{(0)} = w s^{\mathsf{T}},
  \end{equation}
  where $w \in \mathbb{P}^{K_0
  }$ and $s \in \mathbb{P}^T$.
  Additionally, let us introduce matrix $B \in \mathbb{R}^{K_0 \times T}$ with components given by
  \begin{equation}
    \label{eq:B}
    \lbrace B \rbrace_{d,t} := \sum\limits_{k_1 = 1}^{K_1} \lbrace \gamma_{(1)} \rbrace_{k_1,t}
    \left(
      \lbrace X(t) \rbrace_{d} - S_{d,k_1}
    \right)^2.
  \end{equation}
  Then\footnote{It is easy to check that $\epsilon_{0,w} + \epsilon_{0,s} = \epsilon_{0}.$}
  \begin{itemize}
    \item the solution of $w$-problem (where $s$ is fixed) is given by
      \begin{displaymath}
        w^{*} = \softmax \left( - \epsilon_{0,w}^{-1} B s \right), ~~~ \epsilon_{0,w} := \frac{\log(K_0)}{\log(K_0T)} \epsilon_0
      \end{displaymath}
    \item the solution of $s$-problem (where $w$ is fixed) is given by
      \begin{displaymath}
        s^{*} = \softmax \left( - \epsilon_{0,s}^{-1} B^{\mathsf{T}} w  \right), ~~~ \epsilon_{0,s} := \frac{\log(T)}{\log(K_0T)} \epsilon_0.
      \end{displaymath}
  \end{itemize}
  The overall computational complexity of these two steps is $\mathcal{O}(K_1 K_0 T)$.
\end{lemma}

\begin{proof}
  Using the considered factorization \eqref{eq:gamma0_case3}, the entropy regularization of $\gamma_0$ can be written as
  \begin{displaymath}
    \begin{array}{rcl}
      \epsilon_0
      \mathbf{H}(\gamma_{(0)}) & = &
      \frac{\epsilon_0}{\log(K_0T)}
      \sum\limits_{t=1}^{T}
      \sum\limits_{d=1}^{K_0}
      \left\{\gamma_{(0)}\right\}_{d,t}\log(\left\{\gamma_{(0)}\right\}_{d,t})
      =
      \frac{\epsilon_0}{\log(K_0T)}
      \sum\limits_{t=1}^{T}
      \sum\limits_{d=1}^{K_0}
      \left\{ w \right\}_{d} \left\{ s \right\}_{t}
      \log(\left\{ w \right\}_{d} \left\{ s \right\}_{t}) \\
      & = &
      \frac{\epsilon_0}{\log(K_0T)}
      \sum\limits_{t=1}^{T}
      \sum\limits_{d=1}^{K_0}
      \left\{ w \right\}_{d} \left\{ s \right\}_{t}
      \left(
        \log(\left\{ w \right\}_{d}) + \log(\left\{ s \right\}_{t})
      \right) \\
      & = &
      \frac{\epsilon_0}{\log(K_0T)}
      \left(
        \sum\limits_{t=1}^{T}
        \sum\limits_{d=1}^{K_0}
        \left\{ w \right\}_{d} \left\{ s \right\}_{t}
        \log(\left\{ w \right\}_{d})
      \right)
      +
      \frac{\epsilon_0}{\log(K_0T)}
      \left(
        \sum\limits_{t=1}^{T}
        \sum\limits_{d=1}^{K_0}
        \left\{ w \right\}_{d} \left\{ s \right\}_{t}
        \log(\left\{ s \right\}_{t})
      \right) \\
      & = &
      \frac{\epsilon_0 \log(K_0)}{\log(K_0T)}
      \left(
        \sum\limits_{t=1}^{T}
        \left\{ s \right\}_{t}
        \mathbf{H}(w)
      \right)
      +
      \frac{\epsilon_0 \log(T)}{\log(K_0T)}
      \left(
        \sum\limits_{d=1}^{K_0}
        \left\{ w \right\}_{d}
        \mathbf{H}(s)
      \right)
      = \epsilon_{0,w} \mathbf{H}(w) + \epsilon_{0,s} \mathbf{H}(s).
    \end{array}
  \end{displaymath}
  Since the rest of the objective function is linear in variables $w$ and $s$,
  the solution for $w^{*}$ and $s^{*}$ can be easily derived using Lemma \ref{lemma:LP_with_entropy}.

  The computational complexity depends on the assembly of $B \in \mathbb{R}^{K_0 \times T}$ with the sum of $K_1$ terms, the multiplication with given vector, and the evaluation of the softmax function.
\end{proof}

\begin{lemma}[Coordinate descent solution for general probability matrix $\gamma_{(0)}$]
  \label{lemma:gamma0_case4}
  Let $S, \theta, \gamma_{(1)}, \dots, \gamma_{(N+1)}$ in the problem \eqref{eq:EON_fun_E2} be fixed. We consider a general unknown probability matrix $\gamma_{0} \in \mathbb{R}^{K_0 \times T}$.
  Let matrix $B \in \mathbb{R}^{K_0 \times T}$ be given by \eqref{eq:B}.
  Then the solution of the problem is given by
  \begin{displaymath}
    \mathrm{vec} ({\gamma_{(0)}^{*}}) = \softmax \left( - \epsilon_{0}^{-1}
    \mathrm{vec} (B) \right).
  \end{displaymath}
  The overall computational complexity of this solution is $\mathcal{O}(K_1 K_0 T)$.
\end{lemma}

\begin{proof}
  The key step is to write the first term of objective function \eqref{eq:EON_fun_E2} in vectorized form. Then this term is linear in the variable $\gamma_{(0)}$ and the Lemma \ref{lemma:LP_with_entropy} can be directly applied.

  The computational complexity depends on the assembly of $B \in \mathbb{R}^{K_0 \times T}$ and the evaluation of the softmax function.
\end{proof}

\begin{lemma}
  If $N=1$,  $\gamma_{0} = \frac{1}{T} W \onebb^\top_T $, $\epsilon_0 = \epsilon_W$,  $\epsilon_1 = 0$, $\epsilon_2 = 0$, $\delta_1 = \epsilon_{CL}$, then the EON learning problem with Euclidean distance \ref{eq:EON_fun_E2} is equivalent to the entropic Scalable Probabilistic Approximation (eSPA) \cite{horenkoScalableEntropicBreaching2020a, vecchiESPAScalableEntropyOptimal2022a}.
\end{lemma}

\begin{proof}
  It is straightforward to observe that the parameter $\theta$ of the EON problem is equivalent to the parameter $\Lambda^\top$ of the eSPA problem. Moreover, setting $\epsilon_1=0$ switches off the entropic regularization term for $\gamma_{(1)}$. Then, since $\gamma_{(0)}= \frac{1}{T} W \onebb^\top_T $, the optimization problem \eqref{eq:EON_fun_E2} becomes equivalent to equation (2.3) in \cite{vecchiESPAScalableEntropyOptimal2022a}
\end{proof}

\subsubsection{Solution of the EON problem}
In the following, we show  that the solution of the EON problem (\ref{eq:EON_H_con})  can be performed numerically, with an algorithm that leads to a monotonic decrease of the loss function value, has the iteration complexity $\mathcal{O}(T\sum_{n=0}^{N}K_nK_{n+1})$, and relying on the well-defined (and computable in polynomial time) spectral conditions for choosing the remaining hyperparameters $\left(\epsilon,\delta\right)$ - in such a way, that these choices ensure the existence and uniqueness of the label predictions $\pi(s)$ when applying the trained EON model  (\ref{eq:EON_fun}-\ref{eq:EON_fun_con2}) to unlabeled data $X(s)$ (see Theorem \ref{theorem:eon_test}). Aside from these conditions for the $\left(\epsilon,\delta\right)$-dependent existence and uniqueness of classifier solutions, one can further use the direct relationships between the local unit entropies $H$ from (\ref{eq:EON_H_con}) and the regularization parameters $\epsilon$ for selecting the optimal hyper-parameter values. Fine-tuned selection of the hyperparameters (as well as determination of the optimal values for unit dimensions $\left(K_0,\dots,K_{N+1}\right)$) can be achieved through common hyperparameter search strategies, for instance using exhaustive search over a hyperparameter grid or Bayesian hyperparameter optimization strategies \cite{wuHyperparameterOptimizationMachine2019a}.

In general, for any given test point, according to Theorem \ref{theorem:eon_test}, a solution for the $\gamma_{(0)}$ problem exists and is unique. The obtained value of the new point will reflect the probability that the data point belongs to the  training domain of the model. Thus, performing inference on any new instance using EON will provide both a label probability distribution as well as a confidence level on the features (the decision function and the input reliability function $\gamma_0$, respectively, see Figures~3B and 3C for an example). In another words, the input reliability function $\gamma_0$ provides a measure on how much the new data that has to be evaluated is similar to the training data examples. Such measures of input reliability are not available in the common tools, that provide only the decision function.

\begin{theorem}\label{theorem:eon_train}
  (the convergence and computational complexity of EON model using Euclidean distance)
  For any given set of training data $X_t \in \mathbb{R}^{K_0}, \pi_t \in \mathbb{P}^M, t = 1,\dots,T$ and regularization parameters $\epsilon$ and $\delta$, the EON optimization problem \eqref{eq:EON_fun_E2} is well defined and can be solved with iteration complexity $\mathcal{O}(TK_0K_1 + TK_\textrm{all})$, where $K_\textrm{all}$ is defined in \eqref{eq:EON_solLambda_Kall}. The optimization procedure (Coordinate descent algorithm) is monotonic and leads to a solution.
\end{theorem}
\begin{proof}
  When coordinate descent \ref{alg:subspace} is employed, it leads to monotonic decrease of the objective function as proven in Lemma \ref{lemma:generalsubspace}.
  Furthermore, it must be noted that:
  \begin{itemize}
    \item The $S$-problem has analytic solution and the evaluation complexity is $\mathcal{O}(TK_0K1)$, as shown in Lemma \ref{lemma:Sproblem}.
    \item The $\theta$-problem has analytic solution with evaluation complexity $\mathcal{O}(T K_{\textrm{all}})$, as shown in Lemma \ref{lemma:thetaproblem}.
    \item Each $\Gamma$-problem has always a solution (see Lemma \ref{lemma:Gamma_solvability}), which is unique, for specific choice of regularization parameters (see Lemma \ref{lemma:Gamma_unique}).
    \item The entire $\Gamma$ problem can be solved using coordinate descent algorithm, which solves for individual $\gamma$ layer-by-layer (see Lemma \ref{lemma:Gamma_subspace_algorithm}). Each problem has analytic solution (see Lemma \ref{lemma:LP_with_entropy} and \eqref{eq:layerbylayer_softmax}) and the complexity of evaluation is $\mathcal{O}(T K_{\textrm{all}})$
    \item The above points generalize to different formulations of $\gamma_{(0)}$. The first case does not require solving for $\gamma_{(0)}$, as it is fixed. The second case is equivalent to the eSPA problem. The remaining two cases are described in Lemmas \ref{lemma:gamma0_case3} and \ref{lemma:gamma0_case4}.
  \end{itemize}
  Therefore, in each step the iteration complexity of the procedure, is $\mathcal{O}(TK_0K_1 + TK_\textrm{all})$.
\end{proof}

\begin{theorem}\label{theorem:eon_test}
  (the existence and uniqueness of the solution of the EON problem on unlabeled data)
  For any given data point $X \in \mathbb{R}^{K_0}$ without a matching label, regularization parameters $\epsilon$ and $\delta$, and fixed parameters $S$ and $\theta$ previously learned, there exists a unique solution $\Gamma$, and the iteration complexity is $\mathcal{O}(K_{\textrm{all}})$
\end{theorem}
\begin{proof}
  The problem when an unlabeled data point is provided is similar to the training problem, with the following differences: (i) $S$ and $\theta$ are fixed, and (ii) $\gamma_{(N+1)}$ is unknown.
  In order to find a solution, it is only necessary to solve the entire $\Gamma$ problem, including the last layer (see Lemma \ref{lemma:LP_with_entropy}).
  Thus, the solution exists as shown in Lemma \ref{lemma:Gamma_solvability}, is unique for specific values of the regularization parameters as shown in \ref{lemma:Gamma_unique}, and can be found using coordinate descent iteration, according to \ref{lemma:Gamma_subspace_algorithm},

  In the case of the application of the learned model to new data point $X$, the new $\Gamma$-problem has to be solved, the iteration complexity is $\mathcal{O}(K_{\textrm{all}})$.
\end{proof}

\subsection{Adversarial attacks on EON models}
The mathematical formulation of EON allows a simple way to find "adversarial" points, which are points that lead to the most uncertain outcome of the classification task.
Obtaining such points can be extremely useful because, for example, if the data is originating from a process that can be sampled, those points will be the most informative candidates to sample. If, instead, the system cannot be sampled and the data is a fixed set of features and labels, the model can be originally trained with only a subset of the available data, and refined by the inclusion of adversarial points, thus incrementally adding only training data that will be maximally useful for the procedure, until addition of more points does not result in improvements in validation performance, leading to models trained with minimal amounts of data.

The values of $X^a$ and $\Gamma^a = [ \gamma_{(1)}^a, \dots, \gamma_{(N)}^a]$ can be found as those that minimise the following objective function:

\begin{equation}\label{eqn:advers}
  \begin{array}{rcl}
    L_a(\Gamma, X) & = &
    \sum\limits_{k_1=1}^{K_1} \lbrace \gamma_{(1)} \rbrace_{k_1}
    \sum\limits_{d=1}^{K_0} \lbrace \gamma_{(0)} \rbrace_{d} ~ \big( \lbrace X \rbrace_{d} - S_{d,k_1} \big)^2 -
    \sum\limits_{n=1}^{N}
    \delta_n
    \left(
      \sum\limits_{k_n = 1}^{K_n} \sum\limits_{k_{n+1} = 1}^{K_{n+1}}
      \lbrace \gamma_{(n)} \rbrace_{k_n}
      \lbrace \gamma_{(n+1)} \rbrace_{k_{n+1}}
      \log (\theta^{(n)}_{k_n,k_{n+1}})
    \right)
    \\
    & &

    ~~~~~~~~ +

    \sum\limits_{n=0}^{N}
    \epsilon_n \left( \sum\limits_{k_n=1}^{K_n} \lbrace \gamma_{(n)} \rbrace_{k_n} \log \lbrace \gamma_{(n)} \rbrace_{k_n}
    \right)

  \end{array}
\end{equation}

with fixed $\theta, S, \gamma_{(N+1)}$, under the same constraints as \eqref{eq:EON_fun_E2}.

To find such adversarial points, the following two-step procedure can be employed, using coordinate descent iteration in a similar fashion to the procedure performed during model training:
\begin{itemize}
  \item Find $\Gamma^a$, which results in a $\gamma_{(N+1)}$ that is maximum entropy, while keeping $X^a$ fixed.
  \item Find $X^a$ as the set of points that will be encoded by the model as $\Gamma^a$.
\end{itemize}

\begin{algorithm}
  \caption{Coordinate descent algorithm for finding adversarial points}\label{alg:subspace_advers}
  \KwData{$S, \theta, \gamma_{(0)}$ from a fitted model}
  \KwResult{$\Gamma^a \approx \Gamma^{\iit},X^a \approx X^\iit$}
  \vspace{2mm}
  $\textrm{choose}~ \Gamma^0 \in \Omega_{\Gamma}, X^0 \in \mathbb{R}^{K_0}$ \Comment*[r]{initial approximation}
  $\iit \gets 0$ \Comment*[r]{iteration counter}
  \vspace{2mm}
  \While{the stopping criteria is not satisfied}{
    \vspace{2mm}
    $\Gamma^{\iit+1} \gets \arg \min\limits_{\Gamma \in \Omega_{\Gamma}} L_a(\Gamma, X^\iit)$ \Comment*[r]{solve $\Gamma$-problem}
    $X^{\iit+1} \gets \arg \min\limits_{X \in \mathbb{R}^{K_0}} L_a(\Gamma^{\iit+1},X)$ \Comment*[r]{solve $X$-problem}
    $\iit \gets \iit + 1$\;
  }
  \vspace{2mm}
\end{algorithm}

\paragraph{Finding $\gamma^a$}
The problem of finding $\gamma^a$ is equivalent to the solution of the $\Gamma$ step during the training phase described in \ref{par:sol_gamma}, with the only difference being $\gamma_{N+1}$ is in this case considered to be fixed (more specifically to the maximum entropy distribution).

\paragraph{Finding $X^a$}
Once $\Gamma^a$ is found, using the previous step, $X^a$ can be obtained by multiplying $\gamma_{(1)}$ and $S$, as shown in the following lemma.

\begin{lemma}
  The $d$-th component of the minimizer of \eqref{eqn:advers} with fixed $\Gamma^a$ is given by:
  \begin{equation}\label{eq:adver_sol_x}
    \{X^a\}_d = \sum_{k_1=1}^{K_1} \{ \gamma^a_{(1)}\}_{k_1} S_{d,k_1}
  \end{equation}
\end{lemma}
\begin{proof}
  From the first order necessary optimality condition of the unconstrained problem:
  \begin{displaymath}
    \frac{\partial L_a}{\partial \{X\}_{\hat{d}}} =
    2\sum_{k_1=1}^{K_1}
    \{ \gamma^a_{(1)}\}_{k_1}
    \{ \gamma^a_{(0)}\}_{\hat{d}}
    \left( \{ X \}_{\hat{d}} - S_{\hat{d},k_1} \right) = 0.
  \end{displaymath}
  Noting that $\sum_{k_1=1}^{K_1} \{ \gamma_{(1)}\}_{k_1} = 1$ and assuming that $\{\gamma_{(0)}\}_{\hat{d}} \neq 0$, the above condition can be simplified to \eqref{eq:adver_sol_x}.

\end{proof}

\subsection{Numerical experiments}\label{sec:numexp}
In order to describe the performance of the model introduced in the previous section, we tested it on a set of synthetic examples, described in detail in Section \ref{sec:synthetic}. The shared idea behind those examples is that their complexity can be precisely quantified, and this enables a direct comparison between the complexity of the task, in terms of Kolmogorov complexity of the problem, that of the data, i.e., the amount of storage needed to encode each entry in the dataset and that of the learned model. This way, the proposed algorithm can be compared to state-of-the-art tools for the analysis of tabular data in terms of classification performance, computational cost of fitting and evaluating the model, as well as the complexity of the obtained model.

\subsubsection{Application to synthetic examples}

\begin{figure}[h]
  \centering
  \includegraphics[width=1\linewidth]{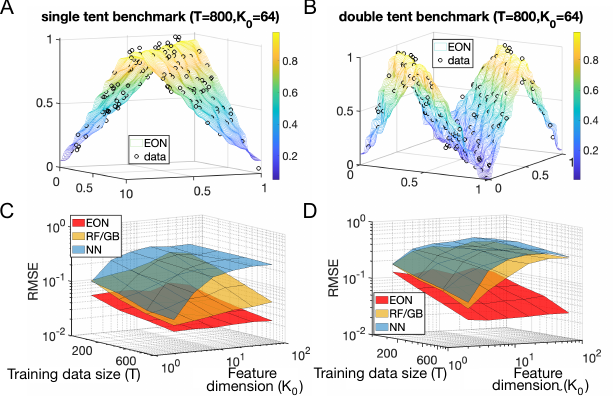}
  \caption{Benchmark results for two regression learning tasks on synthetic examples. (A) Depiction of the single tent benchmark. Data points are shown as points in their first two dimensions (the only informative ones), and the output of EON is shown as a surface, evaluated over a grid. (B) Depiction of the double tent benchmark. The synthetic example is illustrated as in (A). (C) Performance (Root Mean Squared Error, RMSE) for EON and the rest of the compared models (best of Random Forest/Gradient Boosting and Neural Networks) on the single tent benchmark for varying problem dimensionality and number of instances in the training set. (D) Performance (Root Mean Squared Error, RMSE) for EON and the rest of the compared models (best of Random Forest/Gradient Boosting and Neural Networks) on the double tent benchmark for varying problem dimensionality and number of instances in the training set.
  }\label{fig:regression}
\end{figure}

\paragraph{Regression benchmarks}
To investigate the performance of the EON model for regression, we used two synthetic examples, which are scalable in both the number of training samples and the feature dimensionality.
Figure~\ref{fig:regression}A shows the projection onto the first two dimensions for the simpler example, while Figure~\ref{fig:regression}B displays the same projection for the more complex case. The variable to be predicted is shown in the $z$-axis.
In each figure only the first two dimensions are shown, but in both cases the dataset contains 62 additional features that are uniformly distributed and unrelated to the dependent variable, therefore irrelevant for the regression task.

The output of the best-performing EON model is overlaid on both panels, illustrating how closely the prediction follow the true target variable.
Quantitative results for a range of dimensionalities and training set sizes are presented in Figures \ref{fig:regression}C (simple example) and \ref{fig:regression}D (complex example).
In every tested scenarios, the performance achieved by EON (measured using the Root Mean Square Error, RMSE) is consistently lower than that of both the best neural network (NN) baseline and the best ensemble model (RF/GB, selected to be the best perfoming between Random Forest and Gradient Boosted Decision Trees).
The figure illustrates how EON can outperform the other methods despite scarcity of training data as well as in high-dimensional regimes. Both alternative methods exhibit signs of decreasing performance, especially with few high-dimensional data points available for training (i.e., in the small-data regime).

\begin{figure}[h!]
  \centering
  \includegraphics[width=.9\linewidth]{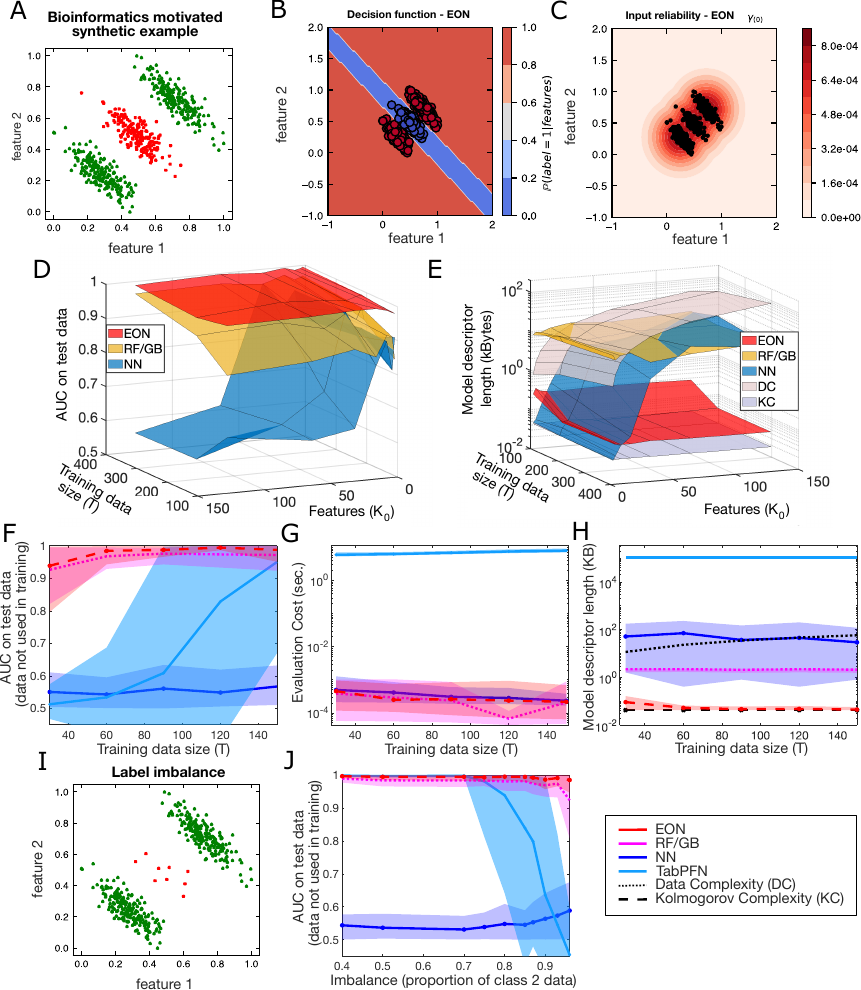}
  \caption{Benchmark results on the bioinformatics motivated synthetic dataset. (A) Projection of the bioinformatics motivated dataset on the first two dimensions. (B) Decision function for the best EON model. (C) Reliability of the input ($\gamma_{(0)}$) for the same EON model. (D). (E). DC refers to the complexity of the data, while KC refers to the Kolmogorov complexity (see the text for precise definition). (F-H) Comparison of the AUC (F), the prediction cost (G) and the model complexity (H, indicated by mean minimal descriptor length) of the benchmarked models as a function of the size of the data used for training. (I) Illustration of label imbalance in the dataset. (J) Comparison of the AUC obtained on the test data while varying the label imbalance in the training dataset.}
  \label{fig:synthetic_a}
\end{figure}

\paragraph{Bioinformatics motivated synthetic example}
In order to showcase the usefulness of the input reliability measure $\gamma_{(0)}$, we provide an example applying the model on a classification example inspired by bioinformatics, described in \cite{horenkoScalableEntropicBreaching2020a} and shown in Figure \ref{fig:synthetic_a}A. As for the regression example, only the first two dimensions (which are shown) are relevant for the classification task, whereas the remaining 4 dimensions are not informative.
Since only the first two dimensions of this synthetic learning example contain information on how to solve the classification problem, in order to successfully address the learning task, the models must be able to identify which particular combination of features should be taken into account - and which feature dimensions are non-informative for a given learning task and have to be discarded.

The simple intrinsic rule that provides the exact classification in this problem requires a 6-dimensional vector for distinguishing which two dimensions are informative (with 6 parameters), together with two lines in two informative dimensions that separate the three distributions (4 parameters), and three labels to encode the class affiliation for each of the three distributions (3 parameters). It means that the exact Kolmogorov complexity,  defined as $KC(K,D)=\sup_{X}\inf_M\text{card}_{\theta_M}\left[M(\theta_M |X)\right]$, where $\text{card}_{\theta_M}$ denotes the cardinality with respect to the parameters $\theta_M$ (i.e., the minimal sufficient number of  parameters required to encode the model $M$) is equal to 13 sufficient parameters  \cite{wallaceMinimumMessageLength1999}
The optimal EON model obtained in training has $N=1$ and $K_1=3$, its descriptor length is 15 parameters. It is closely matching the $KC$ of the original problem, as the parameters needed to perform the EON classifications are: the locations of the (three) centroids, only in the first two dimensions, one row of the $2\times 3$ stochastic matrix $\theta$, and the  $6$-dimensional vector of feature probabilities. The learned rules can be examined by inspecting the decision functions of the best model obtained in the crossvalidation (in Figure \ref{fig:synthetic_a}B).

Note that the evaluation grid for the decision function extends beyond the hypercube containing the original data, and no information on how to classify points in this region is contained in the training dataset. A typical ML model, despite having never observed data in this region during training, still provides predictions for test points outside the "known" region. However, EON offers an additional advantage, thanks to $\gamma_{(0)}$: it provides a reliability score indicating how far each point on the evaluated grid is from the training data, measured in the dimensions that were found to be relevant for this problem (see Figure~\ref{fig:synthetic_a}C). This quantity provides the model's confidence in its classification of the provided and yet unlabeled input, reflecting whether similar data points were encountered during training or not.

Another advantage of using EON is its robustness with respect to both training set size and dimensionality of the features, as illustrated in Figure~\ref{fig:synthetic_a}D and Figure~\ref{fig:synthetic_a}E.
Across all conditions, EON exhibits a remarkably higher tolerance to data scarcity, and is able to operate in the small data regime, with only few high-dimensional data point. Simultaneously, the number of parameters required remains very low and close to the Kolmogorov Complexity (KC) of the underlying problem whereas competing models typically require order of magnitude more parameters, similar to the amount of data used for training (Data Complexity, DC).

A more extreme condition of data scarcity is shown in Figure~\ref{fig:synthetic_a} panels F through H. In this comparison, we additionally included TabPFN, a recently introduced transformer-based foundational model for tabular analyis of small data problems \cite{hollmann2025accurate}.
This model was not included in the previous grid-search due to its significantly higher computational cost.
In this comparison, we focus on varying only the training data size, down to only 30 samples being available for training the model.
The results indicate how, even under those extreme conditions, the obtained EON models are consistently performant (Figure~\ref{fig:synthetic_a}F), fast (Figure~\ref{fig:synthetic_a}G) and close to the Kolmogorov Complexity of the problem (Figure~\ref{fig:synthetic_a}H).

\begin{figure}[h]
  \centering
  \includegraphics[width=.9\linewidth]{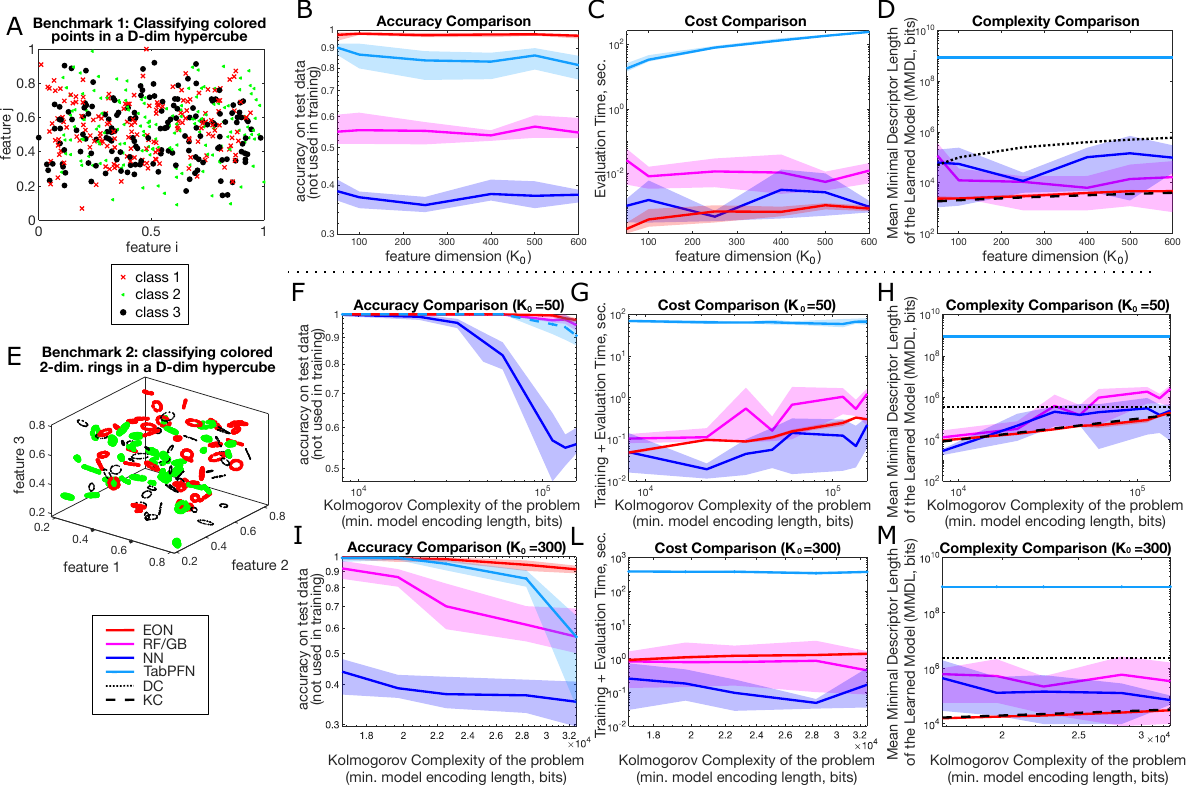}
  \caption{Benchmark results on the synthetic datasets.   (A) Projection of the stacked Gaussians dataset onto dimensions $i$ and $j$. Symbols indicate the class of each data point. (B-D) Comparison of the test accuracy (B), prediction cost (C) and model complexity (D) of the benchmarked methods for the stacked Gaussians datasets with increasing dimensionality.  (E) Projection of the circles dataset onto the first three dimensions. Symbols indicate the class of each data point. (F-H) Comparison of the accuracy (F), the prediction cost (G) and the model complexity (H) of the benchmarked methods for increasing Kolmogorov complexity of the problem. (I-M) are analogous to (F-H) but with dimensionality of the problem set to 300. DC refers to the complexity of the data, while KC refers to the Kolmogorov complexity (see the text for precise definition).}
  \label{fig:synthetic_b}
\end{figure}

Moreover, the performance of EON for classification is also remarkably resilient to class imbalance, which is a common occurrence in real-world datasets.
In this benchmark, we systematically varied the ratio between the two classes, to capture the capability of the used models to perform under these conditions.
Figure~\ref{fig:synthetic_b}D shows how EON is able to achieve near perfect performance even in the case where only 3\% of data points used for training belong to the minority class, whereas the competing methods exhibit performance degradation in highly imbalanced settings.

\paragraph{Benchmark 1}
Next, we present two more extensive sets of benchmarks, in which the scaling properties are investigated more in detail. The first benchmark is a multiclass classification problem on a stack of multivariate Gaussian distributions randomly rotated in 10 dimensions (Figure~\ref{fig:synthetic_b}A). As shown in Figure~\ref{fig:synthetic_b}B, different models present heterogeneous performance in this task. Neural Networks and Random Forest/Gradient Boosting models, in particular, are not able to generalize to the test data, and display overall low accuracy.
TabPFN and EON are the only methods that provide models able to achieve good generalization on this task, with EON consistently maintaining accuracy values close to 1 across all tested dimensionalities of the dataset.

However, the cost of running the inference is quite different across different models. As can be seen in Figure~\ref{fig:synthetic_b}C, EON models are more cost-effective while TabPFN requires a substantial amount of computation, which scales with  the dimensionality of the data and is consistently more than one order of magnitude higher than the other models. RF/GB and NN display intermediate cost, which is paired with inferior performance.
Finally, Figure~\ref{fig:synthetic_b}D illustrates the complexity of the models obtained with each method, in terms of the size of their parameters. EON models are consistently on the boundary of the Kolmogorov Complexity (KC) of the problem, while NN and RF/GB are closer to the data complexity. It is worth noting that the size of TabPFN model is constant, as it is a pretrained network and is applied for all complexities.

\paragraph{Benchmark 2}
Similar results can be observed in the second synthetic benchmark, consisting of a collection of points distributed along randomly rotated rings in a $D$-dimensional hypercube (Figure~\ref{fig:synthetic_b}E).
We illustrate in Figure~\ref{fig:synthetic_b}(F-H) the results obtained on a lower dimensionality (D=50) version of the problem and in Figure~\ref{fig:synthetic_b}(I-M) those on a higher dimensionality one (D=300).
In this example, all methods effectively solve simple problems, especially in the lower dimensionality case (D=50). However, NNs show signs of struggle, as illustrated by the strong decrease in accuracy under more challenging conditions. EON consistently outperforms competitors by producing models that closely approach the Kolmogorov Complexity of the tasks. Moreover, EON's training and evaluation times remain competitive with other methods.

When the problem involves larger dimensionality, as shown in Figure~\ref{fig:synthetic_b}I-M, the situation changes significantly. Under these conditions, NNs fail to generalize their performance to the test data. Similarly, RF/GB models also exhibit poorer outcomes compared to the lower dimensionality setting, though to a lesser extent. The training and evaluation time for EON remains comparable to that of NNs and RF/GB models. However, EON maintains its accuracy even for more complex problems and approaches closely the Kolmogorov complexity of the problems. TabPFN shows a decrease in performance only under the most extreme conditions, but with higher cost and model size.

\subsubsection{Benchmark on open datasets}
In order to investigate whether the promising results obtained on the synthetic datasets extends to real life situations, the performance of EON was evaluated on a diverse set of publicly available datasets spanning several realistic domains of application, ranging from medical diagnostics to physical system simulations.
Table~\ref{tab:classification_perf} reports the median area under the receiver‑operator curve (AUC) for each classification task, while Table~\ref{tab:classification_size} lists the corresponding model sizes together with a comparison to the data size (i.e., the number of entries in the data matrix, reported as the data points multiplied by the number of features per sample).  For regression, Tables~\ref{tab:regression_perf} and \ref{tab:regression_size} present the median root‑mean‑square error (RMSE) and the number of parameters in the best‑performing model for each problem.

\begin{table}[h!]
  \centering
  \begin{tabular}{|c|c|c|c|}
    \hline
    Data                                                          & EON                         & RF/GB         &  NN           \\
    \hline
    mouse behavior \cite{mice_protein_expression_342}             &\textbf{0.997 [0.005]}       &0.996 [0.007]              &0.982 [0.018] \\
    diabetes \cite{turney_pima}                                   &\textbf{0.827 [0.024]}       &0.811 [0.051]              &0.727 [0.036] \\
    lung cancer \cite{mr__abhinaba_biswas_mr__akash_nath_2024}    &\textbf{0.964 [0.008]}       &0.958 [0.020]              &0.928 [0.009] \\
    lung auscultation \cite{lung_auscultation}                    &0.994 [0.000]                &\textbf{0.995 [0.002]}     &0.994 [0.001] \\
    biological human age \cite{age_pred}                          &\textbf{0.747 [0.054]}       &0.685 [0.190]              &0.670 [0.032] \\
    heart failure \cite{heart_failure_519}                        &\textbf{0.880 [0.081]}       &0.870 [0.073]              &0.717 [0.104] \\
    leukemia \cite{leukemia}                                      &0.997 [0.013]                &\textbf{1.000 [0.002]}     &0.990 [0.112] \\
    \hline
  \end{tabular}
  \caption{Median area under the ROC curve (AUC) achieved by EON, RF/GB and NN across a suite of classification problems. The interquartile range is shown in brackets; the best‑performing model for each task is highlighted in bold. For a complete description of the datasets, see the Methods section.}
  \label{tab:classification_perf}

  \centering
  \begin{tabular}{|c|c|c|c|c|}
    \hline
    Data                                                          & EON                           & RF/GB                             & NN                            & Data Size\\
    \hline
    mouse behavior \cite{mice_protein_expression_342}             &\sn{1.29}{4} [\sn{2.38}{3}]     & \sn{1.25}{4} [\sn{2.55}{5}]       & \sn{8.68}{2} [\sn{4.30}{2}]   & \sn{4.16}{4}\\
    diabetes  \cite{turney_pima}                                  &\sn{2.50}{1} [\sn{3}{0}]        & \sn{5.46}{2} [\sn{1.30}{3}]       & \sn{1.79}{3} [\sn{5.50}{2}]   & \sn{4.30}{3}\\
    lung cancer  \cite{mr__abhinaba_biswas_mr__akash_nath_2024}   &\sn{3.85}{2} [\sn{3.19}{1}]     & \sn{1.49}{3} [\sn{2.01}{3}]       & \sn{1.51}{4} [\sn{2.00}{4}]   & \sn{1.22}{4}\\
    lung auscultation \cite{lung_auscultation}                    &\sn{1.30}{1} [0]                & \sn{4.48}{3} [\sn{9.00}{3}]       & \sn{2.40}{3} [\sn{3.93}{4}]   & \sn{1.80}{4}\\
    biological human age \cite{age_pred}                          &\sn{4.30}{1} [\sn{1.70}{1}]     & \sn{5.98}{2} [\sn{8.18}{3}]       & \sn{7.80}{3} [\sn{4.00}{4}]   & \sn{1.28}{4}\\
    heart failure \cite{heart_failure_519}                        &\sn{4.90}{1} [\sn{8.00}{0}]     & \sn{8.22}{2} [\sn{5.67}{2}]       & \sn{1.82}{2} [\sn{3.00}{1}]   & \sn{2.51}{3}\\
    leukemia \cite{leukemia}                                      &\sn{6.11}{2} [0]                & \sn{4.29}{2} [\sn{9.27}{2}]       & \sn{9.35}{1} [\sn{7.59}{2}]   & \sn{1.26}{4}\\
    \hline
  \end{tabular}
  \caption{Median number of parameters in the best performing EON, RF/GB, and NN models for the classification problems in Table \ref{tab:classification_perf}. The interquartile range is indicated in brackets.}
  \label{tab:classification_size}
\end{table}

\paragraph{{Classification problems}}
Across five out of seven classification tasks, EON outperforms all of the considered standard methods, both in terms of higher median AUC performance and lower model complexity (see Tables 1 and 2). For two out of seven benchmarks (lung auscultation and leukemia) RF/GB has outperformed EON in median AUC, however, the advantage in both of the case is within the interquartile confidence intervals, and, hence, not very significant.
As for the synthetic examples, the advantage of using EON is not limited to its predictive accuracy.
Indeed, Table~\ref{tab:classification_size} shows that EON models are consistently more compact.
The median parameter count is roughly an order of magnitude lower than that of RF/GB and two orders of magnitude below the neural‑network baseline. Moreover, comparing the model sizes to the data size values reveals that EON is the only method that reliably delivers models whose parameter count is lower than the dimensionality of the training dataset. The only two benchmarks where EON was outperformed by the existent methods (RF/GB) in terms of the median AUC were auscultation and leukemia datasets \cite{lung_auscultation,leukemia}. However, as can be seen from the Tables 1 and 2, EON exhibits a much smaller interquartile ranges for AUC and model descriptor length - and hence, it provides much more compact and robust models. Compactness of EON models makes them much easier to analyse and interpret: for example, for the auscultation benchmark - with training data size of 10'800 floating point numbers - EON requires only 13 tunable parameters, whereas the best standard model (RF/GB) requires on average 4'480 tunable model parameters for this benchmark.

\paragraph{{Regression problems}}
Table~\ref{tab:regression_perf} demonstrates that EON achieves the lowest RMSE on all of the six considered regression tasks.
As for the classification benchmarks and in contrast to the standard methods, EON models can attain such performance while using not only significantly less tunable parameters, but much less than the size of the training datasets.
In a few instances (for example, the prediction of SOD1 gene expression, or for the classification problem on the same data set) the NN models exhibit fewer parameters than EON.  However, this apparent size advantage for NN is offset by a markedly worse predictive performance and RMSE values than those obtained with EON.
Please note that for the algometry and tetany tasks, it is also possible to perform predictions using pharmacokinetic/pharmacodynamic model (PK/PD) based on first principles from physics and chemistry , which achieves RMSE of 0.058 [0.007] and 0.046 [0.008], respectively, with only 5 tunable model parameters \cite{kern2004response} - that represents an estimate of the Kolmogorov Complexity bound for this problem. As can be seen from Tables 3 and 4, EON achieves basically the same level of precision (RMSE of 0.056) and complexity (21 to 25 tunable model parameters).  

\begin{table}[h!]
  \centering
  \begin{tabular}{|l|c|c|c|}
    \hline
    Data                                                        & EON                       & RF/GB         & NN            \\
    \hline
    anesthetics (algometry) \cite{kern2004response}             &\textbf{0.070 [0.016]}     &0.082 [0.009]  &0.106 [0.015]  \\
    anesthetics (tetany) \cite{kern2004response}                &\textbf{0.056 [0.010]}     &0.065 [0.014]  &0.091 [0.022]  \\
    Lorenz-96, type 1, F=4 \cite{lorenz1996predictability}      &\textbf{0.011 [0.002]}     &0.019 [0.026]  &0.185 [0.070]  \\
    Lorenz-96, type 1, F=8 \cite{lorenz1996predictability}      &\textbf{0.018 [0.003]}     &0.030 [0.022]  &0.265 [0.022]  \\
    liver  \cite{liver_uci}                                     &\textbf{0.152 [0.026]}     &0.156 [0.025]  &0.204 [0.026]  \\
    SOD1 expression  \cite{mice_protein_expression_342}         &\textbf{0.022 [0.004]}     &0.041 [0.008]  &0.086 [0.031]  \\
    \hline
  \end{tabular}
  \caption{Median root-mean-square error (RMSE) achieved by EON, RF/GB, and NN across a collection of regression tasks. The interquartile range is shown in brackets; the best‑performing model for each task is highlighted in bold. For details on the datasets, see the Methods section. }
  \label{tab:regression_perf}

  \centering
  \begin{tabular}{|c|c|c|c|c|}
    \hline
    Data                                                        & EON                           & RF/GB                             & NN                            & Data Size            \\
    \hline
    anesthetics (algometry) \cite{kern2004response}             &\sn{2.50}{1}[\sn{6.00}{0}]     & \sn{9.87}{3} [\sn{1.17}{4}]       & \sn{4.04}{4} [\sn{2.98}{4}]   & \sn{6.36}{2}  \\
    anesthetics (tetany) \cite{kern2004response}                &\sn{2.10}{1}[\sn{8.00}{0}]     & \sn{1.78}{4} [\sn{4.59}{4}]       & \sn{4.55}{4} [\sn{4.36}{4}]   & \sn{6.36}{2}  \\
    Lorenz-96, type 1, F=4 \cite{lorenz1996predictability}      &\sn{1.14}{3}[\sn{1.85}{2}]     & \sn{5.62}{3} [\sn{7.62}{3}]       & \sn{1.90}{4} [\sn{1.57}{4}]   & \sn{1.61}{3}  \\
    Lorenz-96, type 1, F=8 \cite{lorenz1996predictability}      &\sn{9.39}{2}[\sn{1.80}{2}]     & \sn{6.74}{3} [\sn{1.19}{4}]       & \sn{1.39}{4} [\sn{1.46}{3}]   & \sn{1.62}{3}  \\
    liver  \cite{liver_uci}                                     &\sn{1.90}{1}[0]                & \sn{3.83}{3} [\sn{1.36}{4}]       & \sn{1.24}{4} [0]              & \sn{1.66}{3}  \\
    SOD1 expression  \cite{mice_protein_expression_342}         &\sn{3.04}{4}[\sn{2.40}{3}]     & \sn{2.49}{4} [\sn{4.94}{4}]       & \sn{9.63}{3} [\sn{4.05}{3}]   & \sn{3.93}{4}\\

    \hline
  \end{tabular}
  \caption{Median number of parameters in the best performing EON, RF/GB, and NN models, for the regression tasks listed in Table \ref{tab:regression_perf}. The interquartile range is indicated in brackets.}
  \label{tab:regression_size}
\end{table}

\subsubsection{Application to a climate prediction problem}\label{sec:real_world}
Finally, we consider the problem of data-driven predictions of the El Niño and La Niña phases of the El Niño Southern Oscillation (ENSO), a climate phenomena governing the interannual variation of sea surface temperatures in the Pacific Ocean. The problem was selected because the periodicity of ENSO is highly irregular, and its prediction has recently been gaining growing importance due to its recognized major impact on global climate and economy \cite{mcphadenENSOIntegratingConcept2006}. Consequently, obtaining accurate predictions of the NINO3.4 index using modern AI tools is currently a task receiving significant research efforts \cite{zhaoExplainableNinoPredictability2024}. In the experiments presented below, we consider the problem of predicting the phase of ENSO, i.e., transforming the learning task into a classification problem, and transforming the time series into i.i.d. data using delay embedding. The feature values $X$ used for training in this example represent a series of anomalies with respect to the climatological monthly averages for the dominant processes known to excite intraseasonal variability in the tropical Pacific ocean consisting of the 100 dominant Empirical Orthogonal Function projections (EOFs) of global sea surface temperature and the 100 dominant deep ocean water vertical temperature gradient EOFs in the equatorial Pacific Ocean. A detailed explanation of this data set is provided in \cite{vecchiESPAScalableEntropyOptimal2022a}.

\begin{figure}[H]
  \centering
  \includegraphics[width=.9\linewidth]{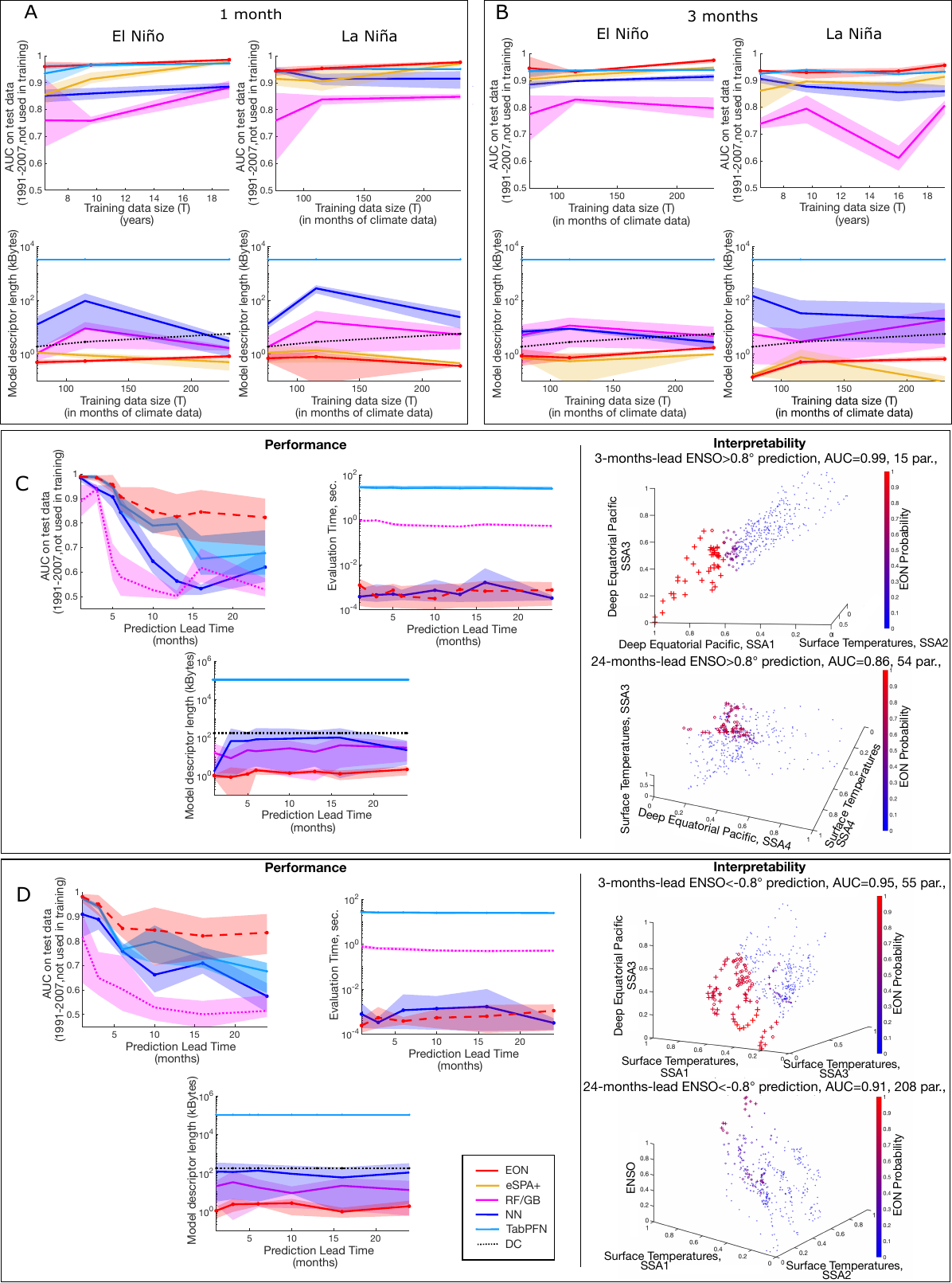}
  \caption{Results on ENSO prediction. (A) Top panels represent the performance (AUC) on test set, for El Niño (left) and La Niña (right) with 1 month lead time, bottom panels illustrate the model descriptor length. (B) As in panel A, but with 3 months lead time.
  (C) Prediction of extreme El Niño events for varying prediction lead times. The models were trained using data from 1975 up to 1990, the plots illustrate AUC, model descriptor length and training and evaluation cost. On the right panels, it is possible to observe the projection of the data and the respective labels on the most significant dimensions identified by the models. (D) is equivalent to (C), but for prediction of extreme La Niña events.}
  \label{fig:enso}
\end{figure}

For the benchmark, we performed prediction of the phase of ENSO, for both El Niño and La Niña (as a separate binary classification problem) on data from the period 1991 to 2007. The models were trained on data collected between 1975 and 1994.
To avoid data contamination (i.e., using the same data for training and testing the models), when the training data extended after 1991, the test dataset was reduced accordingly.
The confidence intervals are obtained by sampling from the neighborhood of the point defining the end of the training data period. The prediction is performed at the lead time of 1 and 3 months ahead, while varying the amount of data available for training the models, the Area Under the ROC Curve (AUC) is illustrated in Figure~\ref{fig:enso}A. The performance of the models can be seen to generally improve when using more training data. Notably, EON consistently achieves elevated AUC values throughout the considered period, for both ENSO phases, surpassing NNs and RFs/GBs. TabPFN also demonstrates promising results for this task. However, its performance comes with the drawback of significantly larger model size, as shown in Figure~\ref{fig:enso}B. In contrast, the models obtained using entropic learning (EON and eSPA+) remain the most compact for both phases of ENSO and both lead times.

Two major challenges in ENSO prediction are (i) performing forecasts at longer lead times, and (ii) learning extreme and rare events, imposing the class imbalance challenge on the learning task. Figure~\ref{fig:enso}C and D report the results for increasing prediction horizon, using binary labels that indicate whether the ENSO index exceeded 0.8° or fell below -0.8° (corresponding, respectively, to extreme El Niño and La Niña events). The panels show the performance (AUC), the size of the best model and the cost of training plus evaluation. Across all horizons, EON models achieve AUC values that are comparable to or surpass those obtained by other state of the art methodologies, while simultaneously exhibiting smaller size and reduced computational cost.

Moreover, in the right sides of Figure~\ref{fig:enso}C and Figure~\ref{fig:enso}D illustrate the level of introspection that is possible to obtain when using EON, as the most relevant dimensions can be readily plotted in order to inspect what feature the model is considers important. Note how in the cases shown projecting onto the specific combination of dimensions reveals separable boundary between the classes.

\section{Methods}\label{sec:methods}
The algorithms considered in the benchmark include commonly used tools for learning from tabular data: Neural Networks (NN), Random Forests (RF), Gradient Boosting (GB), and TabPFN \cite{hollmann2025accurate}.

\subsection{Bioinformatics motivated synthetic example}
This example is generated as described previously \cite{horenkoScalableEntropicBreaching2020a}, using a total of 6 dimensions and 600 data points. Among the dimensions, two are informative with respect to the classification, while the remaining four contain random values drawn from a uniform distribution.
A 20-Fold crossvalidation strategy was employed, with hyperparameters tuned using a single nested shuffle split with test size of 50 points.

The following hyperparameter choices were explored: $K \in \{ 3,6,9\}$, $\delta \in { 10^{-3}, 10^{-2}, 10^{-1}}$, $\epsilon_1 \in \{ 10^{-8}, 5 \times 10^{-8}, 5\times 10^{-7}\}$. For each combination, 10 different initializations were fitted, and the best performing model was retained.
$\gamma_{(0)}$ was built as the outer product of stochastic vectors: the regularization for the entropy of dimensions was selected between $\{3\times10^{-3}, 3\times10^{-1}\}$, while that of data points between $\{ 10^{-3},10^{-2}, 5\times 10^{-2}, 10^{-1}, 5\times10^{-1} \}$.

\subsection{Hyperparameter selection}
Performance evaluation and hyperparameter selection for benchmarks 1 and 2 were performed using 50 fold Monte Carlo crossvalidation. For each fold, the data was split into three non-overlapping subsets: train, validation and test. Exhaustive grid search was employed, where for each combination of hyperparameters, a model was trained on the training set. The reported test performance refers to that of the model having highest performance on the validation set.

The hyperparameter used for fitting EON models are listed in the Table \ref{tab:hyp_eon}. Only $N=1$ was employed in the comparisons.
For the regression task, the hyperparameters were randomly sampled in the following intervals: $\delta_1$: $[0.9, 1.1]$, $\epsilon_0$: $[5e-4, 5e-2]$, $\epsilon_1$: $[1e-3, 5e-2]$, $K$: $[\lfloor\frac{3}{9} T\rfloor, \lfloor \frac{8}{9}T \rfloor]$.
Neural Networks (NN) were fitted using the 'fitcnet' function in MATLAB, with the optimal 'LayerSize' hyperparameter searched using the following list of possible values: 1, 2, 3, 4, 5, 10, 12, 15, 25, 30, 50, 70, 100, 125, 150, 175, 200, 225, 250, 275, 300, 350, 400, 450, [5,5], [10,10], [20,20], [50,50], [100,100], [150 150], [200 200], [5 5 5], [10 10 10], [25 25 25],[100,100,100], and [150,150,150].
Random Forest models were trained using the 'fitcensemble' function with Bayesian hyperparameter optimization used with its default settings, except for fixing the 'Method' to 'Bag'.
Gradient Boosted Decision Trees (GB) models were trained using the 'fitcensemble' function, but with trees as base learners.
For the regression task, Random Forest and Gradient Boosted Decision Trees models were trained as for the classification task, with the difference that the function used was 'fitrensemble'. Similarly, for Neural Networks the same configuration was used, but using 'fitcnet'.
All experiments were conducted in MATLAB R2024a, except for TabPFN, which was run using the Python package available at \url{https://github.com/PriorLabs/TabPFN}, without hyperparameter tuning (since foundational models do not require hyperparameter tuning). To ensure a fair comparison of the computational cost, all methods were evaluated on the CPU.
All compared algorithms were provided with the same training, validation and testing data in all of the crossvalidation splits, and run on the same hardware: a Macbook Pro with an M1 Max processor and 64GB of RAM.

\begin{table}[h]
\centering
\begin{tabular}{|c|c|}
\hline
  Hyperparameter & Values\\
  \hline
  $\delta$ & 1e-5, 1e-4, 1e-3, 5e-3, 1e-2, 1e-1, 1, 10\\
  $\epsilon_{0}$ & 1e-3, 3e-3, 5e-3, 4e-3, 8e-3\\
  $\epsilon_{1}$ & 1e-12, 1e-6, 1e-5, 5e-5, 1e-4, 5e-4, 1e-3, 5e-3\\
  $K$ & 3,4,5,6,7,8\\
  \hline
\end{tabular}
\caption{Hyperparameters for EON}
\label{tab:hyp_eon}
\end{table}

\subsection{Synthetic datasets}\label{sec:synthetic}
\paragraph{Regression}
The performance of EON on regression learning was probed using two synthetic examples:
\begin{enumerate}
\item A "single tent" benchmark can be created by sampling uniformly distributed points in a $K_0$ dimensional hypercube and for each of them calculate the target variable to be:
  \begin{equation}
    y(t) =
    \begin{cases}
      \begin{pmatrix}
        1\\1
      \end{pmatrix}^\top
      x(t) + \mathcal{N}(0,0.01)~~~~~~~\text{if}~~~\sum_{d=1}^2 \{x(t)\}_d < 1\\
      \begin{pmatrix}
        -1\\-1
      \end{pmatrix}^\top
      (x(t)-1) + \mathcal{N}(0,0.01)~~~~~~~\text{otherwise}
    \end{cases}
  \end{equation}
\item A "double tent benchmark differs in the creation of $y(t)$, which is:
  \begin{equation}
    y(t) = 1 - 2\left(\frac{1}{2} \vert \sum_{d=1}^2 \{x(t)\}_d \vert + \lfloor \sum_{d=1}^2 \{ x(t)\}_d \rceil \right) +\mathcal{N}(0, 0.01)
  \end{equation}
\end{enumerate}
In both cases, $y$ is rescaled to be in the interval $[0,1]$. This setup allows for flexible generation of learning problems with varying number of dimensions and training samples.

\paragraph{Classification}
The synthetic learning problems used to test the performance of classification methods consist of randomly-generated datasets within a $D$-dimensional hypercube:
\begin{enumerate}
\item The first set of test problems is a classification of a stack of linearly-ordered $K$ normal distributions that can be separated with $(K-1)$ distinct linear $D$-dimensional manifolds. An example is illustrated in Figure \ref{fig:synthetic_b}A.
\item The second is the classification of randomly-centered and randomly-oriented two-dimensional rings of random radii. An example is illustrated in Figure~\ref{fig:synthetic_b}E.
\end{enumerate}
For any selection of the hypercube dimensions $D$ and the number of objects $K$, both of these synthetic classification problems allow for the computation of the exact Kolmogorov complexity, which can be defined as $KC(K,D)=\sup_{X}\inf_M\text{card}_{\theta_M}\left[M(\theta_M |X)\right]$, where $\text{card}_{\theta_M}$ denotes the cardinality with respect to the parameters $\theta_M$ (i.e., the minimal sufficient number of bytes required to store the parameters required to encode the model $M$) \cite{wallaceMinimumMessageLength1999}.
Additionally, the worst-case complexity expected from a model can be calculated as $DC(T,K_0) = \sup_{X_{\text{train}}} \text{card}(X_{\text{train}})$, which is equivalent to the size of all data points presented in the training set.

\subsection{Non-synthetic real-life datasets}\label{sec:real}
The open datasets used for classification and regression benchmarks are summarized in Table \ref{tab:datasets}.
They contain a mix of binary (e.g., diabetes) as well as multiclass (e.g., the eight-class mouse behavior task) problems.
All feature vectors were normalized to the interval $[0,1]$ using min-max scaling before being used for learning.
Model evaluation followed the crossvalidation protocol described above. For datasets with less than 1000 data points, the splits used for train/validation/test were 70/15/15, while for the rest 80/10/10 was employed.

\begin{table}[h]
\centering
\begin{tabular}{|l|c|c|}
  \hline
  Name                                                            & Features ($K_0$)  & Number of data instances ($T$) \\
  \hline
  mouse behavior \cite{mice_protein_expression_342}               & 51                & 1080              \\
  diabetes \cite{turney_pima}                                     & 8                 & 768               \\
  lung cancer \cite{mr__abhinaba_biswas_mr__akash_nath_2024}      & 15                & 1157              \\
  lung auscultation \cite{lung_auscultation}                      & 3                 & 12000             \\
  biological human age \cite{age_pred}                            & 8                 & 2278              \\
  heart failure \cite{heart_failure_519}                          & 12                & 299               \\
  leukemia \cite{leukemia}                                        & 14                & 1281              \\
  \hline
  anesthetics (algometry)   \cite{kern2004response}               & 2                 & 397               \\
  anesthetics (tetany)  \cite{kern2004response}                   & 2                 & 397               \\
  Lorenz-96, type 1, F=4 \cite{lorenz1996predictability}          & 400               & 1151              \\
  Lorenz-96, type 1, F=8 \cite{lorenz1996predictability}          & 160               & 1157              \\
  liver  \cite{liver_uci}                                         & 5                 & 345               \\
  SOD1 expression  \cite{mice_protein_expression_342}             & 51                & 1080              \\
  \hline
\end{tabular}
\caption{Description of the datasets used in this work.}
\label{tab:datasets}
\end{table}

\section{Discussion}\label{sec:discussion}
In this work, we introduce EON as a reformulation of Boltzmann Machines (BMs), deploying the probability-conforming distance measures, as well as relaxing the standard equilibrium assumption of BMs. As was shown above, it allows several significant conceptual advantages compared to the state-of-the-art learning tools: it results (i)  in straightforward proofs of existence and uniqueness conditions for problem solutions, (ii) in fast, cheap and gradient-descent free training, and (iii) in robustness measures computable in polynomial time.
Despite of the higher level of mathematical generality (due to the more general assumptions like the non-equilibrium assumption), these mathematical results were much more straightforward to obtain - and in practical applications are computationally demonstrably less expensive than the standard results for percolation/Ising models and BMs \cite{smirnov2010conformal}.

Unlike traditional AI tools, such as neural networks, which rely on backpropagation and gradient descent for learning, EON leverages a sequence of analytical solutions implemented as the dimension descent  algorithm. It does not require any of the standard gradient-descent-based training algorithms, hence eliminating the need for tuning the respective hyperparameters (like learning rate, momentum weights and similar). It was shown that the introduced approach circumvents the limitations associated with neural networks and gradient-based optimization methods, in particular the need for significant amounts of data and computational resources for training: as shown in the Figure~4, EON requires only few (10-18) years of climate data to achieve the predictability plateaus - whereas the standard tools require much more of the training data (see also Figures. 3 and 4).
The training procedure of EON can be carried out, for a given set of hyperparameters tailored to a specific problem, by selecting only the desired tolerance and maximal number of steps allowed.
The efficacy of EON, notably superior rates of convergence and Kolmogorov complexity, relative to NNs, arise due to its foundational mathematical properties enabling analytical solutions at each step of the learning procedure.

As illustrated on the synthetic and real examples, the proposed EON tool provides significant advantages, extending beyond its low complexity (measured as a model descriptor length) and training cost to include easy interpretability of the resulting models, as well as in availability of the input reliability measure $\gamma_{(0)}$ (see Figure~3C). The learnable parameter $\gamma_{(0)}$ can be used to indicate the level of confidence of the model (as well as the importance of each feature) when presented a sample of test data points, as it indicates the proximity of the point to the hull of the training dataset. Moreover, the straightforward mathematical formulation allows for the transparent analysis and understanding of the obtained model \cite{horenkoExistenceUniquenessScalability2023, groom2024comparative}.

\subsection{Key findings and implications}
Formal proofs are provided for all mathematical properties of the proposed method including solutions for both the learning and the evaluation/labeling problems. To validate its effectiveness, we compared EON's performance against a comprehensive set of state-of-the-art methods for classification and regression, on a broad class of synthetic and real examples of varying problem complexity, ranging from synthetic scalable datasets, to challenges appearing in biomedicine, fluid mechanics and climate science. Across all benchmarks, we observed superior performance of the proposed method in key critical performance factors (AUC on test data, model descriptor length, training cost). Specifically, it was shown that EON enables training of more compact models in less time, while achieving better prediction performance. This efficiency can be of particular use in time-sensitive applications and on hardware with limited computational resources.

Moreover, EON addresses the issue of "overconfidence" often found in many state-of-the-art AI tools. Provided mathematical theory and numerical algorithms open a straightforward and cheap possibility to obtain a level of confidence in the predictions, enabling a more reliable, interpretable and humble decision-making process, particularly suitable for applications that require explainable decisions, such as in climate research, healthcare and other sensitive areas.

\subsection{Limitations and future directions}
One direction for further work is in extending this approach to a broader set of AI problems, such as for example generative modeling and agent-based decision making, as well as in reinforcement learning. Moreover, EON could be adapted to handle the unique challenges presented by application domains, such as life sciences, where it is crucial to deal with outliers and imbalanced datasets, as well as with the integration of heterogeneous sources of data such as multi-modal inputs. Further tailoring of the methodology may be necessary to optimize its performance under these special circumstances.

Application examples considered above indicate that EON represents a significant step towards solving small-data learning problems - when dealing with high feature dimensions and relatively small sample data statistics.
Summarizing, we envision that this methodology can help to address a wide range of challenges in different fields.

\section*{Acknowledgments}
This work was funded by the European Commission under Horizons European Programme (Grant Agreement No. 101080756), for the AI4LUNGS consortium, and travel support was provided by MLKL (Machine Learning Consortium in Kaiserslautern). We wish to thank Christoph Schütte (ZIP/FU-Berlin) for the helpful comments about the manuscript.

\section*{Declarations}
\subsection*{Conflict of interest/Competing interests}
The authors declare no conflict of interests.

\subsection*{Data and code availability}
The code reproducing the results presented in this work will be shared upon reasonable request.

\bibliographystyle{unsrt}

\end{document}